\documentclass[runningheads]{llncs}
\usepackage[T1]{fontenc}
\usepackage{graphicx,verbatim}
\usepackage{booktabs}
\usepackage{acronym}
\usepackage{wrapfig}
\usepackage{amsmath}
\usepackage{adjustbox}
\usepackage{amssymb}
\usepackage{amsfonts}
\usepackage{multirow}
\usepackage{float}
\usepackage{tikz}
\usepackage{xcolor}
\usepackage{subcaption}
\usepackage{array}
\usetikzlibrary{bayesnet}
\usepackage[colorlinks,breaklinks]{hyperref} 
\usepackage{xurl} 
\usepackage{hyperref}
\usepackage{cleveref}
\begin{document}
\title{Graph Conditioned Diffusion for Controllable
Histopathology Image Generation}
\titlerunning{Graph Conditioned Diffusion}

\author{Sarah Cechnicka\inst{1} \orcidID{0009-0008-3449-9379}  \and
Matthew Baugh\inst{1}\ \orcidID{0000-0001-6252-7658} \and
Weitong Zhang \inst{1} \orcidID{0000-0002-3681-4546}\and
Mischa Dombrowski \inst{2} \orcidID{0000-0003-1061-8990}\and
Zhe Li \inst{2} \orcidID{0009-0003-3101-718X} \and
Johannes C. Paetzold \inst{3,4} \orcidID{0000-0002-4844-6955}\and
Candice Roufosse \inst{1,5}\orcidID{0000-0002-6490-4290} \and
Bernhard Kainz \inst{1,2}\orcidID{0000-0002-7813-5023}
}
\authorrunning{S Cechnicka et al.}
%
\institute{Department of Computing, Imperial College London, UK \and
Dept. AIBE, Friedrich–Alexander University Erlangen–N\"urnberg, DE \and 
Weill Cornell Medicine, NY, USA \and 
Cornell Tech, NY, USA  \and
Centre for Inflammatory Disease, Imperial College London, London, UK\\ 
\email{sc7718@imperial.ac.uk}\\}

\maketitle              
\begin{abstract}

Recent advances in \acp{dpm} have set new standards in high-quality image synthesis.
Yet, controlled generation remains challenging particularly in sensitive areas such as medical imaging.
Medical images feature inherent structure such as consistent spatial arrangement, shape or texture, all of which are critical for diagnosis. However, existing \acp{dpm} operate in noisy latent spaces that lack semantic structure and strong priors, making it difficult to ensure meaningful control over generated content. 
To address this, we propose graph-based object-level representations for Graph-Conditioned-Diffusion.
Our approach generates graph nodes corresponding to each major structure in the image, encapsulating their individual features and relationships.
These graph representations are processed by a transformer module and integrated into a diffusion model via the text-conditioning mechanism, enabling fine-grained control over generation.
We evaluate this approach using a real-world histopathology use case, demonstrating that our generated data can reliably substitute for annotated patient data in downstream segmentation tasks. The code is available \href{https://github.com/scechnicka/Graph-Conditioned-Diffusion}{here}.


\end{abstract}

\newcommand{\ud}{\mathrm{d}}
\newcommand{\bfx}{\mathbf{x}}
\newcommand{\bff}{\mathbf{f}}
\newcommand{\bfw}{\mathbf{w}}

\acrodef{ml}[ML]{Machine Learning}
\acrodef{wsi}[WSI]{Whole Slide Image}
\acrodef{t5}[T5]{Text-to-Text Transfer Transformer}
\acrodef{clip}[CLIP]{Contrastive Language–Image Pre-training}
\acrodef{vit}[ViT]{Vision Transformer}
\acrodef{ai}[AI]{Artificial Intelligence}
\acrodef{gt}[GT]{Ground Truth}
\acrodef{com}[COM]{Center of Mass}
\acrodef{vae}[VAE]{variational autoencoder}
\acrodef{dpm}[DPM]{Diffusion Probabilistic Model}
\acrodef{gan}[GAN]{Generative Adversarial Network}
\acrodef{gcd}[GCD]{Graph Conditioned Diffusion}
\acrodef{FID}{Fréchet Inception Distance}
\acrodef{IP}{Improved Precision}
\acrodef{IR}{Improved Recall}
\acrodef{cdm}[CDM]{Conditional Diffusion Model}  
\acrodef{dice}[Dice]{Dice Similarity Coefficient}
\acrodef{aji}[AJI]{Aggregated Jaccard Index}
\acrodef{cnn}[CNN]{Convolutional Neural Network}

\section{Introduction}
\label{sec:intro}
Many medical fields are undergoing a transformation, moving from the traditional, manual examination of samples towards a fully digitized paradigm.
In histopathology, this involves the high-resolution digital scanning of histologically prepared tissue samples, which are then processed into \acp{wsi}.
This shift opens the door for a transition from primarily qualitative assessments to more quantifiable, automated, and feature-based analysis using \ac{ml} algorithms.
However, the vast size of \acp{wsi}, with resolutions upwards of 60,000 by 60,000 pixels per sample, containing thousands of structures, makes it a challenging task.
This complexity, coupled with the difficulties in manual annotation for supervised \ac{ml} and medical data sharing constraints~\cite{NASDM}, poses significant barriers to the clinical adoption of \ac{ml} in digital pathology.

\begin{figure}[t]
    \centering
    \includegraphics[width=\columnwidth]{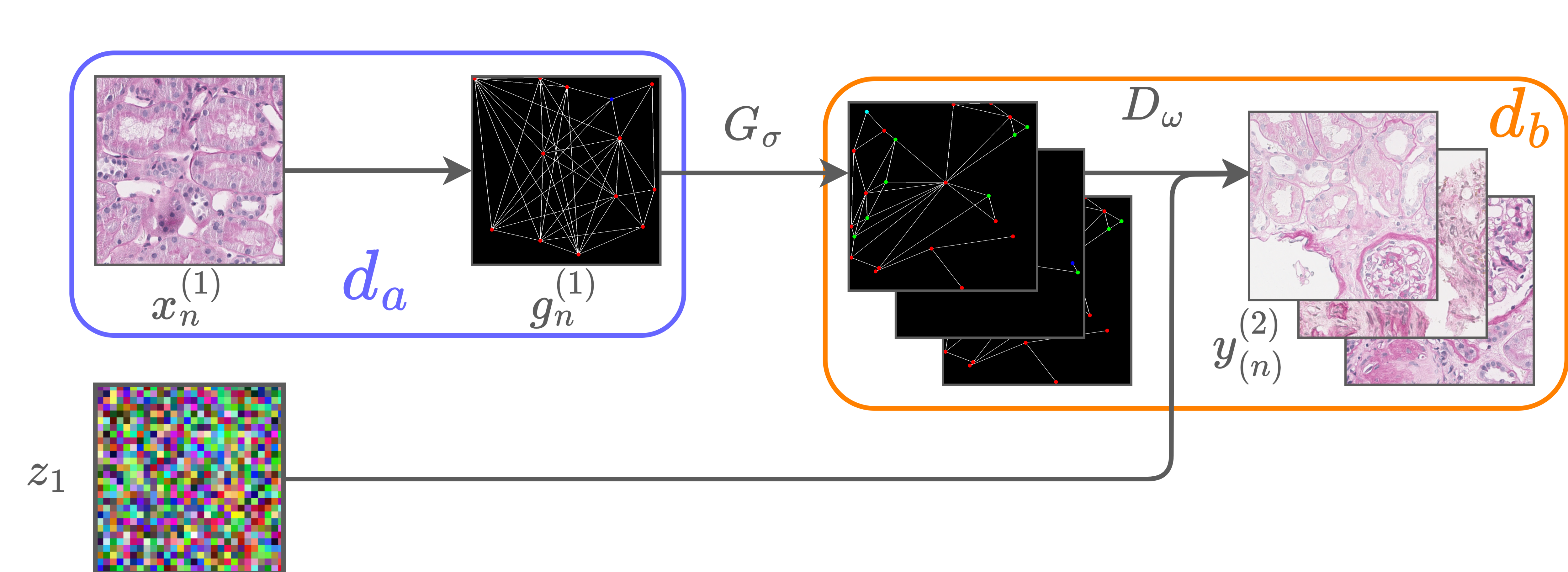}
    \caption{
    Overview of our graph-based controllable synthesis pipeline. Graphs extracted from semantic image information (left) are processed by $G_{\sigma}$(e.g. node removal, node changes, node interpolations) to identify optimal graph representations. These representations condition $D_{\omega}$'s to generate synthetic histopathology images (right), creating datasets with enhanced utility for segmentation and diagnostic applications.}
    \label{fig:teaser}
\end{figure}

There has been a move towards generating synthetic samples to overcome the limitations of insufficient data in the face of privacy issues~\cite{privacy_med}.
Expanding existing datasets with data augmentations has been explored~\cite{FARYNA2024108018}, but fail to generate semantic variety by their nature as image operations.
Techniques such as \acp{gan}~\cite{gan}, \acp{vae}~\cite{vae}, and diffusion models~\cite{NASDM} have shown promising results in learning the underlying image patterns and bridging domain gaps.
However, despite synthetic images becoming increasingly indistinguishable from real ones~\cite{Hadrien,Cechnicka2024}, a shift in the distribution of sample diversity persists.
This discrepancy impacts the efficacy of models trained solely on synthetic data, as they might not accurately represent the wide range of variations found in real-world samples. 
Part of the issue again lies in the data, as the distribution of data produced by a probabilistic generative model is as unbalanced as the data it was trained on.
Without explicit control over feature distributions, the generation process may reinforce existing biases rather than remedy them.
Thus models trained for downstream tasks on datasets enriched by such samples do not automatically have enhanced performance~\cite{Cechnicka2023}.

In this paper we address the diversity limitations of synthetic datasets and present a novel method based on graph proxy representations for conditional image generation. Our approach allows for control over the diversity of generated samples through \ac{gcd}. Our contributions are: \textbf{(1)} We show that a histopathology image can be represented with a proxy graph, which can accurately guide a diffusion process. \textbf{(2)} To enable effective conditioning, we introduce a textual representation of graphs by providing an original architecture that allows for graph tokenization, replacing textual embeddings~\cite{synthetic_inRL}; \textbf{(3)} We take advantage of these insights to investigate the causal relationships between objects in images by subtly altering their graph representations.
\textbf{(4)} We achieve image generation quality en par with other generation methods while providing synthetic data that faithfully represents the training data distribution for downstream tasks such as image segmentation. 
Our method enables us to closely align the synthetic datasets with the statistical properties of real datasets, enhancing the utility of synthetic datasets for diagnostic applications and low-barrier data sharing. Our approach is outlined in Figure~\ref{fig:teaser}.


\noindent\textbf{Related Works:} 
Diffusion models have demonstrated the ability to generate synthetic images more closely aligned with target dataset distributions~\cite{ramesh_zero-shot_2021,ondistillationofguideddiffusionmodels,imagen,zhang2024stability}.
Recent applications of \acp{cdm}~\cite{imagen,NASDM,Cechnicka2023} have significantly improved the integration of multi-scale information in medical imaging, advancing diagnostic precision and enhancing patient care~\cite{Hadrien}. However, research consistently shows that models trained exclusively on synthetic data underperform compared to those trained on real data when applied to downstream tasks~\cite{Cechnicka2023,Gao2023}, limiting their usefulness in practice. 
\noindent In the context of histopathology, especially within cancer research, graphs have become a of interest for navigating the challenges of \ac{wsi} classifications, enabling detailed analysis beyond the capability of direct \ac{wsi} processing~\cite{GraphTransformer,Histograph,Histograph2}. However, the integration of graphs into image synthesis has not been explored yet.

\section{Method}

\noindent\textbf{Synthetic Distribution Gap Analysis:}
Samples drawn from an unconditional diffusion model are not guaranteed to represent the underlying distribution of the original dataset $\tilde{\mathcal{X}}$.
In fact, the diffusion model will mainly draw highly typical samples close to the distribution's mean $\mu({\tilde{\mathcal{X}}})$.
Conditioning of the diffusion model provides more precise control over the sampling process, but this only helps if the conditioning representation allows for targeted decisions to be made to balance datasets.
Otherwise, there is no guarantee that the distribution's full range, including potential edge cases, will be captured.
We hypothesize that graphs constructed from key features of the images in the original datasets provide such means of conditioning, as they offer a structured and interpretable encoding of the data, allowing for explicit decision-making regarding which structures should appear in specific contexts, and enabling more balanced and diverse sampling while maintaining control over the model’s outputs.


\noindent\textbf{Graph Construction:} 
\ac{gt} graphs are generated using a defined protocol. The pixel-level \ac{com} is calculated for all segmented objects of each image
from the \ac{gt} label masks. 
Different classes are marked with distinct labels, and the vertex set is connected according to the following criterion: for vertices \(v_i\) and \(v_j\) with centers of mass \(C_i\) and \(C_j\), an edge \((i,j)\) exists if and only if the line segment
from \(C_i\) to \(C_j\) does not intersect any other labeled region, i.e.,
$$\forall t\in [0,1], (1-t)C_i+tC_j \notin \bigcup_{k \neq i,j} R_k.
$$
Several examples demonstrating this can be seen in the ``images'' row of Figure~\ref{fig:combined_results}.

\noindent\textbf{Graph Textual Embedding:} 
To condition a diffusion model on a graph structure, we introduce a transformer with modified attention masks, specifically designed to replace the standard auto-regressive text-conditioning mechanism. This involves the use of adjacency matrices $\mathbf{A}$ as the basis for constructing attention matrices, effectively embedding graph structural information into the model.
This means our attention mechanism is defined as  
$
\text{softmax}\left(\frac{\mathbf{Q}\mathbf{K}^T}{\sqrt{d_k}} \odot \mathbf{A}\right), 
$ 
where $\mathbf{Q}$ and $\mathbf{K}$ are the query and key matrices of dimensions $N \times d_k$, derived from the graph nodes' feature embeddings, and $d_k$ is the dimensionality of the key vectors, ensuring the scaled dot-product attention incorporates the adjacency matrix $\mathbf{A}$. 
The feature embeddings $\mathbf{F}$ of nodes are designed to replace traditional textual embeddings and are defined as
$
\mathbf{F} = \mathbf{E}_{class}  \oplus \mathbf{E}_{BYOL} \oplus \mathbf{E}_{pos}, 
$ 
where $\oplus$ denotes concatenation.
The first component $\mathbf{E}_{class}$ is a one-hot vector encoding the nodes class.
The second component $\mathbf{E}_{BYOL}$ is the result of applying a binary mask to the original image such that only pixels belonging to the target object are kept, followed by obtaining a vector embedding using a convolutional neural network trained with BYOL~\cite{grill2020bootstrap} on such masked images.
Finally the third component $\mathbf{E}_{pos}$ is the positional encoding generated using a sinusoidal function from vision transformers.
This configuration enables the diffusion model to be conditioned on the graph by embedding rich, graph-structured information into the model's learning process.

\noindent\textbf{Graph Conditioned Diffusion (GCD):}  
Following graph construction, we leverage these graphs $G$ to condition an SDE-based diffusion process.
The forward diffusion process follows the standard stochastic formulation; gradually adding noise to an image $\mathbf{x_0}$ over time $t \in [0,1]$ to produce images with varying degrees of noise $\mathbf{x_t}$.
This process transforms a structured image into pure noise, enabling the learning of a reverse process that can generate realistic images from noise.
For graph-conditioned generation, the backward process aims to reverse the process conditioned on the constructed $G$.
To facilitate this we treat each graph as a high-dimensional entity characterized by $\mathbf{G}
\in \mathbb{R}^{N \times F} \times \mathbb{R}^{N \times N}$, where $N$ represents the number of nodes with $F$ dimensional features, alongside an adjacency matrix $\mathbf{A}$ for edge attributes. 
After incorporating this as additional conditioning for the neural model, the sample during reverse process $\boldsymbol{x}_s$ is defined as:
\begin{equation}
    \boldsymbol{x_s} = \boldsymbol{\tilde{\mu}}_{s|t}(\boldsymbol{x_t}, \boldsymbol{\hat{x}_\theta(x_t}, \lambda_t, \boldsymbol{g})) + \sqrt{(\tilde{\sigma}^2_{s|t})^{1-\gamma}(\sigma^2_{t|s})^\gamma} \epsilon
    \label{eq:sampling_cgsde}, 
\end{equation} 
where $\lambda_t = \log\left[\frac{\alpha_t^2}{\sigma_t^2}\right]$ represents the noise schedules. The entire GCD model adheres to score-matching principles and incorporates a weighted variational lower bound within the graph embedding framework:
\begin{equation}
    \mathbb{E}_{t, \boldsymbol{x}_0 \sim p(\boldsymbol{x}), \boldsymbol{\epsilon} \sim \mathcal{N}(0, \boldsymbol{I}), g}[w(\lambda_t)\|\boldsymbol{\hat{x}}_\theta(\boldsymbol{x}_t, \lambda_t, g) - \boldsymbol{x}\|_2^2].
    \label{eq:score_cgsde}
\end{equation}

We incorporate this graph conditioning into a cascaded diffusion model~\cite{imagen}, initially generating images at a resolution of $64\times 64$ followed by two super-resolution models upsampling it first to $256\times 256$ and then $1024 \times 1024$.
All three models are augmented with our graph conditioning.

\noindent\textbf{Graph Interventions:} 
Following our flexible sampling in Eq.~\ref{eq:sampling_cgsde}, we propose a graph augmentation mechanism. For each graph $G$, this process begins with minimal modifications, such as the removal of a single node, denoted as $G^{-v}$, or the change of a node’s class from an initial class $c$ to a new class $c'$, represented as $G^{v: c \to c'}$. More generally, any modification to a specific node $v$, such as alterations to its features or attributes, can be expressed as $G^{v,\sigma}$, where $\sigma$ captures the nature of the modification.

These targeted interventions form the foundation for more extensive intervention mechanisms as it has been explored  
for molecular modelling~\cite{jensen2019graph} and image augmentation techniques like Cut-Paste~\cite{dwibedi2017cut}. This process begins by identifying segments within graphs by locating lone bridges-edges whose removal results in the creation of subgraphs. These identified subgraphs are then randomly mixed and matched with other subgraphs from different graphs in the training set, emulating a Cut-Paste methodology adapted for graph structures.
For each graph $G$ in the training set, we identify edges $e$ that, when removed, partition $G$ into two distinct subgraphs, $G_1$ and $G_2$. Then we randomly select pairs of subgraphs $(G_i, G_j)$ from the pool of generated subgraphs across the training set and combine the selected pairs to form new graphs $G_{new} = G_i \oplus G_j$, where $\oplus$ denotes the operation of mixing and matching subgraph structures.

In cases where direct mixing and matching are infeasible or result in raph structures that violate the defigned design constraints, we employ a linear interpolation strategy between two graphs, $G_a$ and $G_b$, to synthesize intermediate graph structures. This option serves as a mechanism to generate variations between existing graph structures, enhancing the diversity of the augmented dataset.

\section{Experiments}
\noindent\textbf{Data Preprocessing:} We use an in-house Kidney Transplant Pathology \ac{wsi} dataset (courtesy of Charing Cross Hospital) containing 334 patients, similar in construction to \cite{monkey_dataset,loupy2022thirty}.
Diagnosis of kidney pathology images is based predominantly on the appearance and configuration of structures such as tubules and glomeruli, meaning correctly replicating the distribution of such objects is vital.
We randomly select 6 \acp{wsi} for testing on downstream tasks, and use the remaining to train the diffusion models.
As each \ac{wsi} has a resolution of $40,000 \times 40,000$ we split them into $1024 \times 1024$ pixel sections and discard any non-tissue or empty patches.
The remaining 1654 sections are annotated for kidney cortex features by an expert pathologist and an assistant, of which 68 belong to the test images. 
In order to train each stage of our cascaded diffusion model we make copies of each patch at both $64\times 64$ and $256 \times 256$ resolution, applying simple data augmentations such as shifting and flipping dynamically during training.

\noindent\textbf{Training setup:} All models are trained on a compute node with eight Nvidia A100 GPUs for an average training time of 72 hours.

\noindent\textbf{Metrics:} To evaluate generative models, we employ several complementary metrics.
The \ac{FID} evaluates generated image quality and diversity by comparing two Gaussian distributions $\mathcal{N}(\boldsymbol{\mu}_r, \boldsymbol{\Sigma}_r)$ and $\mathcal{N}(\boldsymbol{\mu}_g, \boldsymbol{\Sigma}_g)$, where $r$ indicates real images and $g$ represents generated samples.
The mean $\boldsymbol{\mu}$ and covariance $\boldsymbol{\Sigma}$ are computed from latent-space feature vectors extracted from both real and generated images.
Though \ac{FID} serves as a well-established metric for evaluating distributional similarities between real and generated data, additional metrics provide deeper insights. \ac{IP} quantifies how well the generated samples correspond to the real data manifold, while \ac{IR} measures how effectively the synthetic data manifold encompasses the real data distribution, indicating the model's coverage of data diversity. Both metrics utilize k-Nearest Neighbors distances to create non-parametric approximations of these data manifolds.
Together, IR and IP provide insights into a generative model's capability to produce both diverse and high-quality samples that accurately reflect the real data characteristics.
We provide additional experiments on a downstream segmentation task that are evaluated on real \ac{wsi}s to verify the utility of our generated images in a setting closely resembling real-world usage, using \ac{dice}~\cite{dice1945measures} and \ac{aji}~\cite{kumar2017dataset} as evaluation metrics.    

\section{Results} 
We compare our \ac{gcd} against other state-of-the-art diffusion-based methods with results shown in Table~\ref{tab:methodcomparison}.
Since the primary focus of this work is on enhancing diffusion models' capabilities through graph-based representations, rather than a strict evaluation of their generative abilities, the comparisons are based on existing diffusion-based approaches.
As such, an unconditional diffusion model and a mask-conditioned diffusion model were chosen (text-conditioned diffusion was not possible as the dataset has no text labels).
Additionally, besides common fidelity metrics (\ac{IP}, 
 \ac{IR}, \ac{FID}) downstream segmentation tasks were performed to better assess the practical utility of the generated images, evaluating not only their visual quality but also their effectiveness in improving performance on task-specific objectives.
Our method achieves higher \ac{IP} and \ac{IR} scores, indicating that \ac{gcd} captures the diversity of the real image distribution more effectively.
Despite this, the images generated with \ac{gcd} have a lower \ac{FID}, although this is not too surprising as previous work has shown that \ac{FID} does not effectively measure image diversity~\cite{dombrowski2024image,Kynkaanniemi2019}.
Looking at the downstream tasks, the models trained on data produced by \ac{gcd} achieve performance on par with a pure diffusion model while providing the added benefit of control over the image content, whereas the masked diffusion approach performs significantly worse.
Figure~\ref{fig:combined_results} left illustrates the effectiveness of the generated synthetic data for downstream segmentation tasks.

\newcommand\x{0.18\textwidth}
\begin{figure*}[h]
  \centering
  \resizebox{\textwidth}{!}{
    \begin{tabular}{l *{5}{c} p{0.5cm} *{5}{c}}
      & \multicolumn{4}{c}{synthetic} & real & & real & \multicolumn{4}{c}{synthetic} \\
      &  &  &  &  &  & & \scriptsize{$G$} & \scriptsize{$G$} & \scriptsize{$G^{-4}$} & \scriptsize{$G^{-11}$} & \scriptsize{$G^{2:1\rightarrow2}$} \\
      \cmidrule(lr){2-5} \cmidrule(lr){6-6} \cmidrule(lr){8-8} \cmidrule(lr){9-12}
      \rotatebox[origin=l]{90}{\scriptsize{graphs}} & 
        \includegraphics[width=\x]{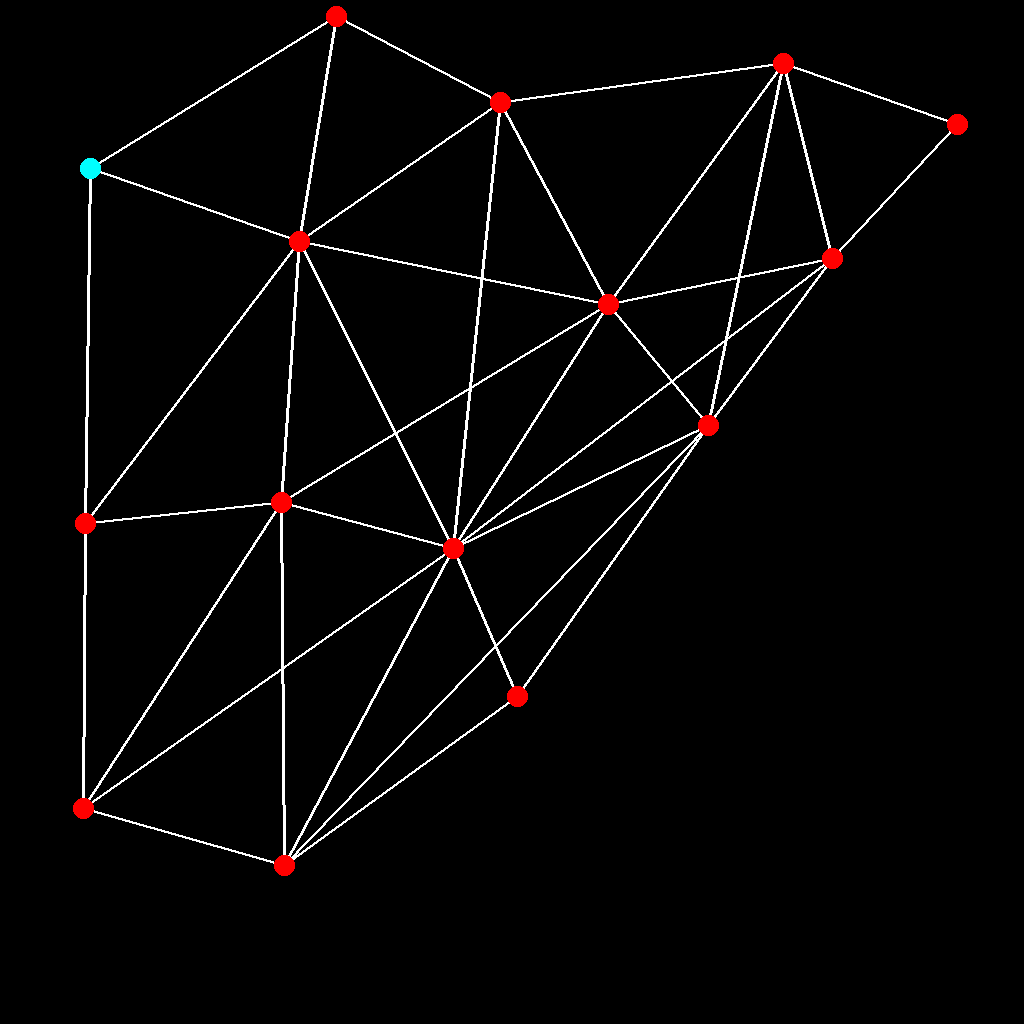} &  
        \includegraphics[width=\x]{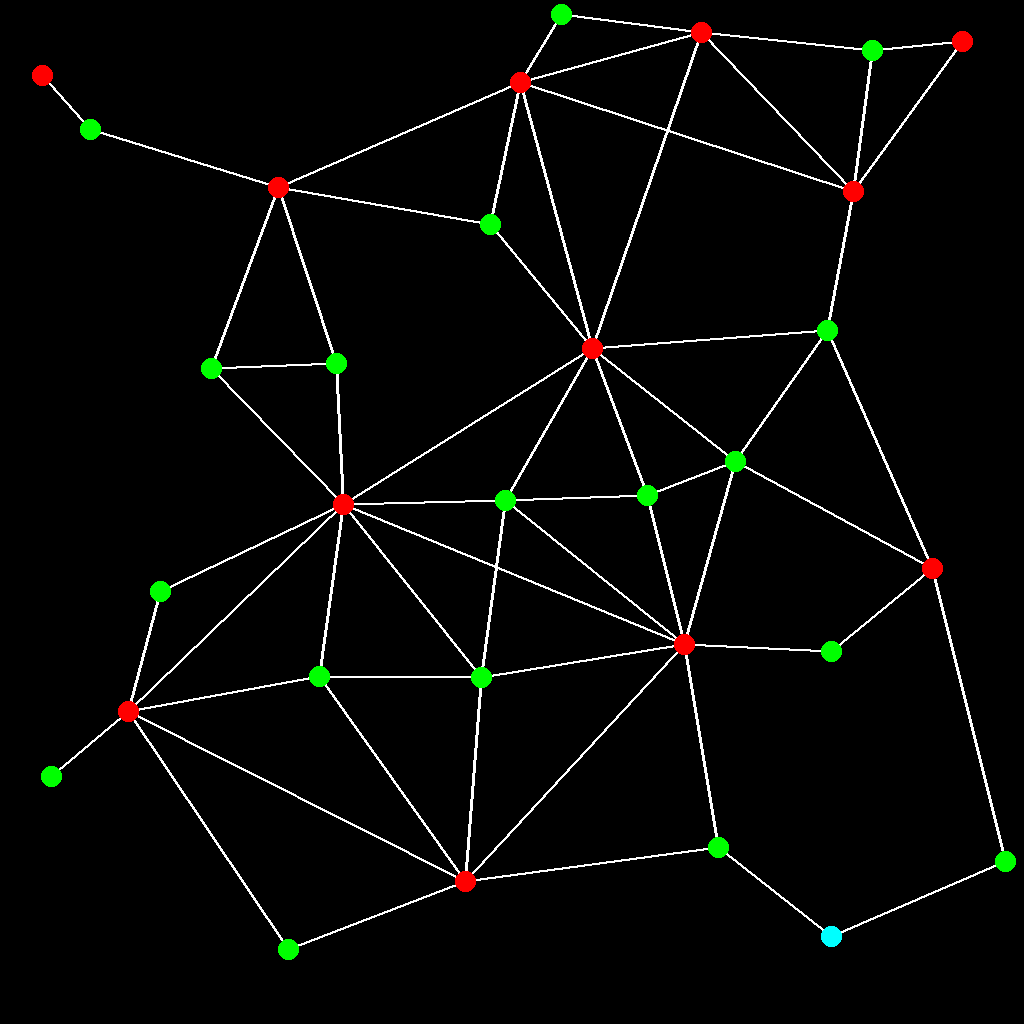} &
        \includegraphics[width=\x]{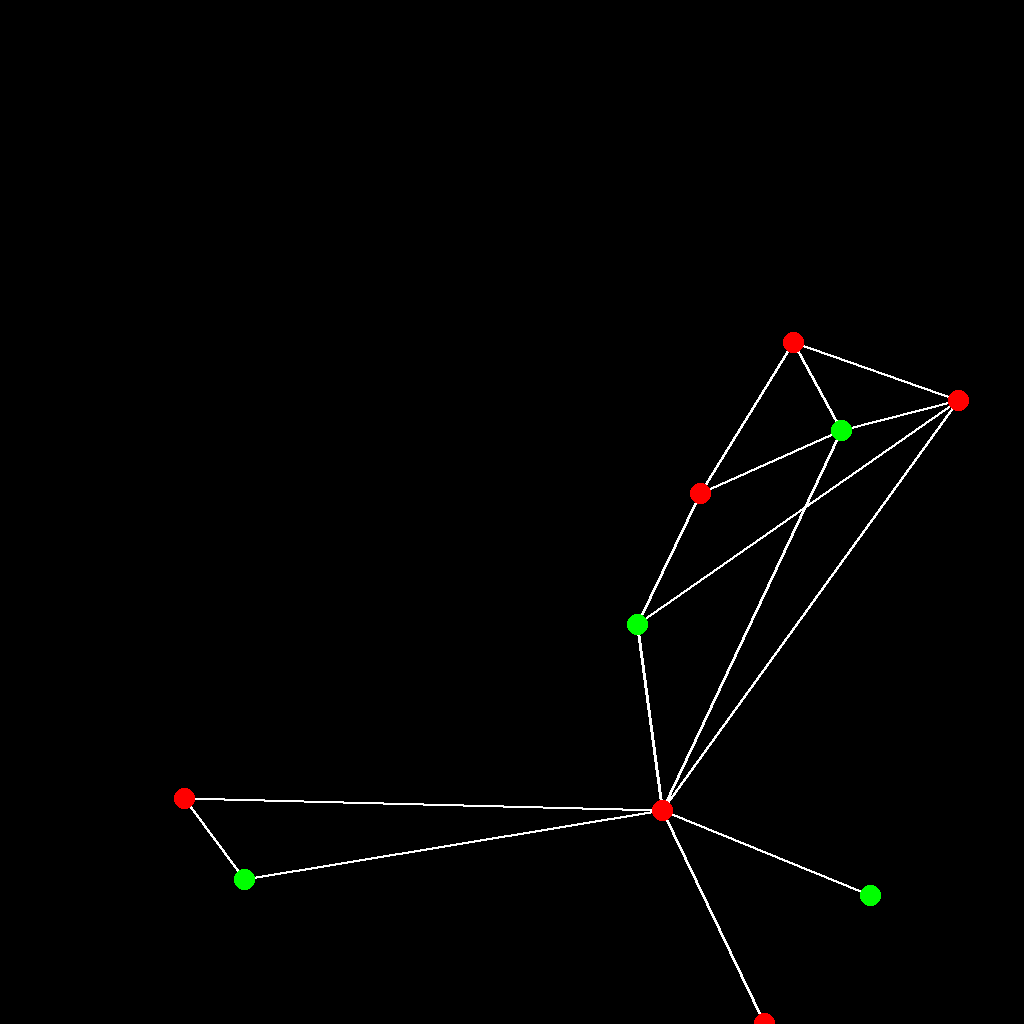} &
        \includegraphics[width=\x]{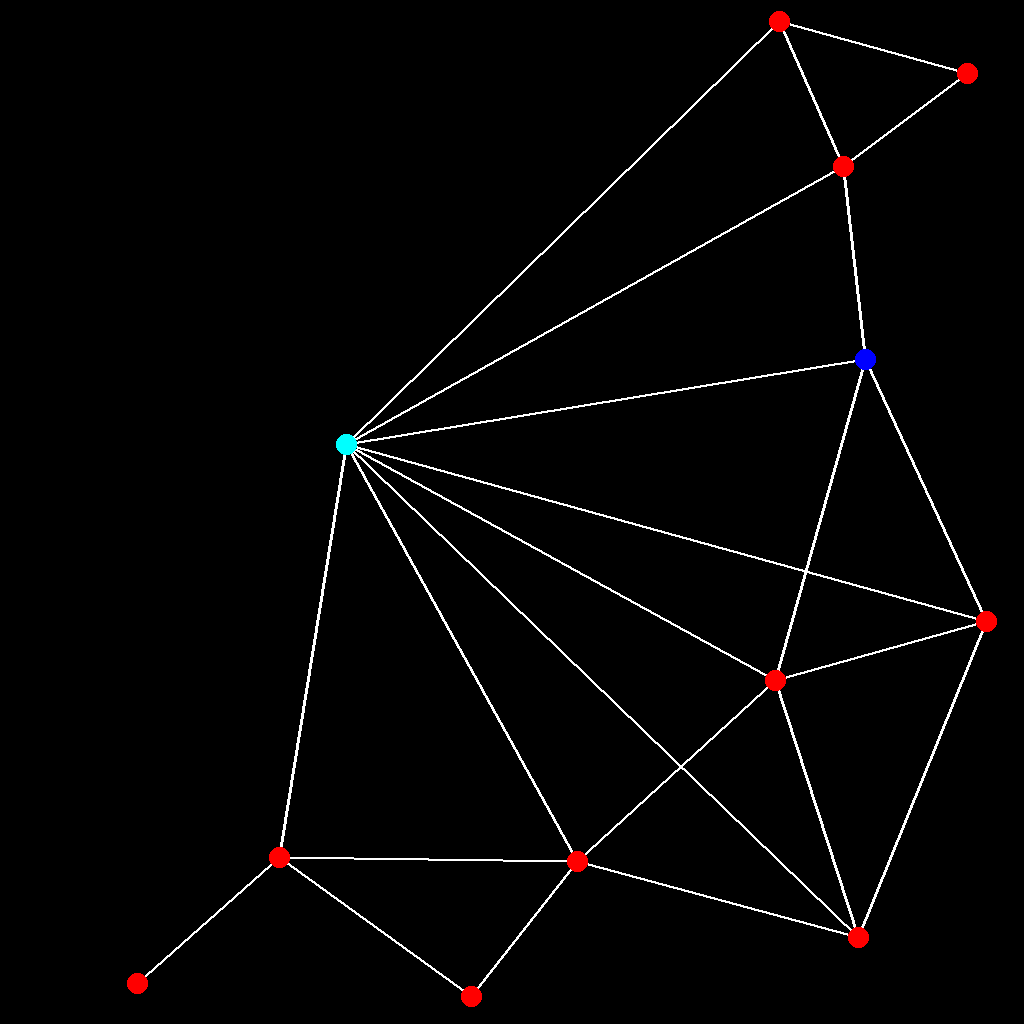} &
        \includegraphics[width=\x]{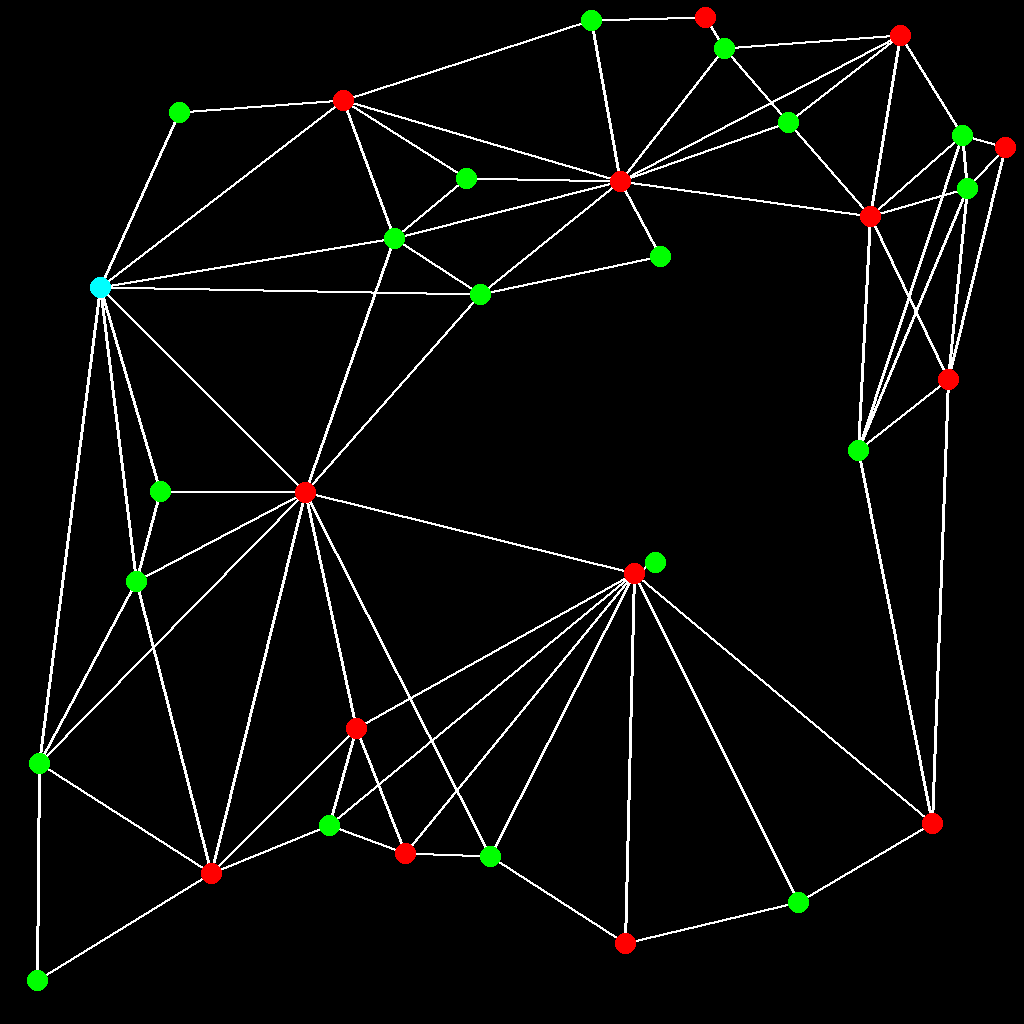} & 
        & 
        \includegraphics[width=\x]{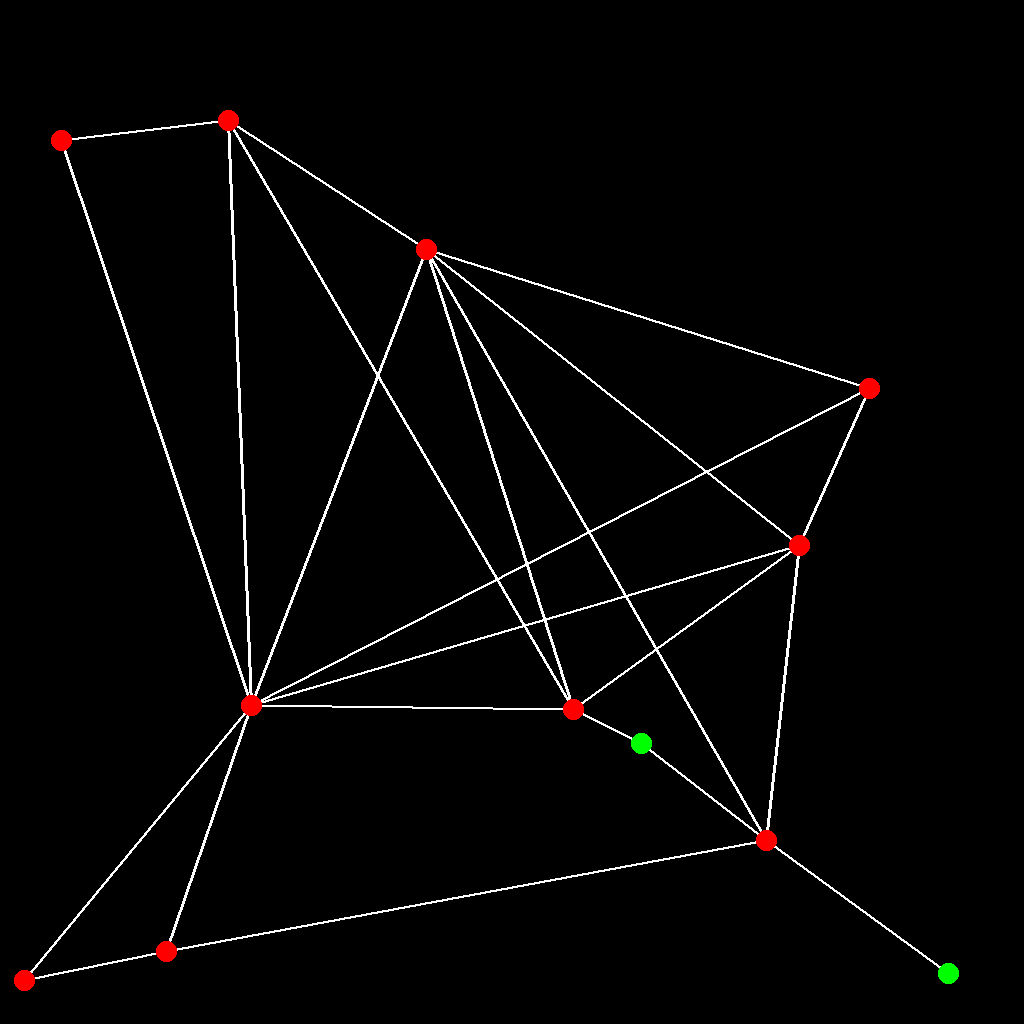} & 
        \includegraphics[width=\x]{images/causal_graph.png} &
        \includegraphics[width=\x]{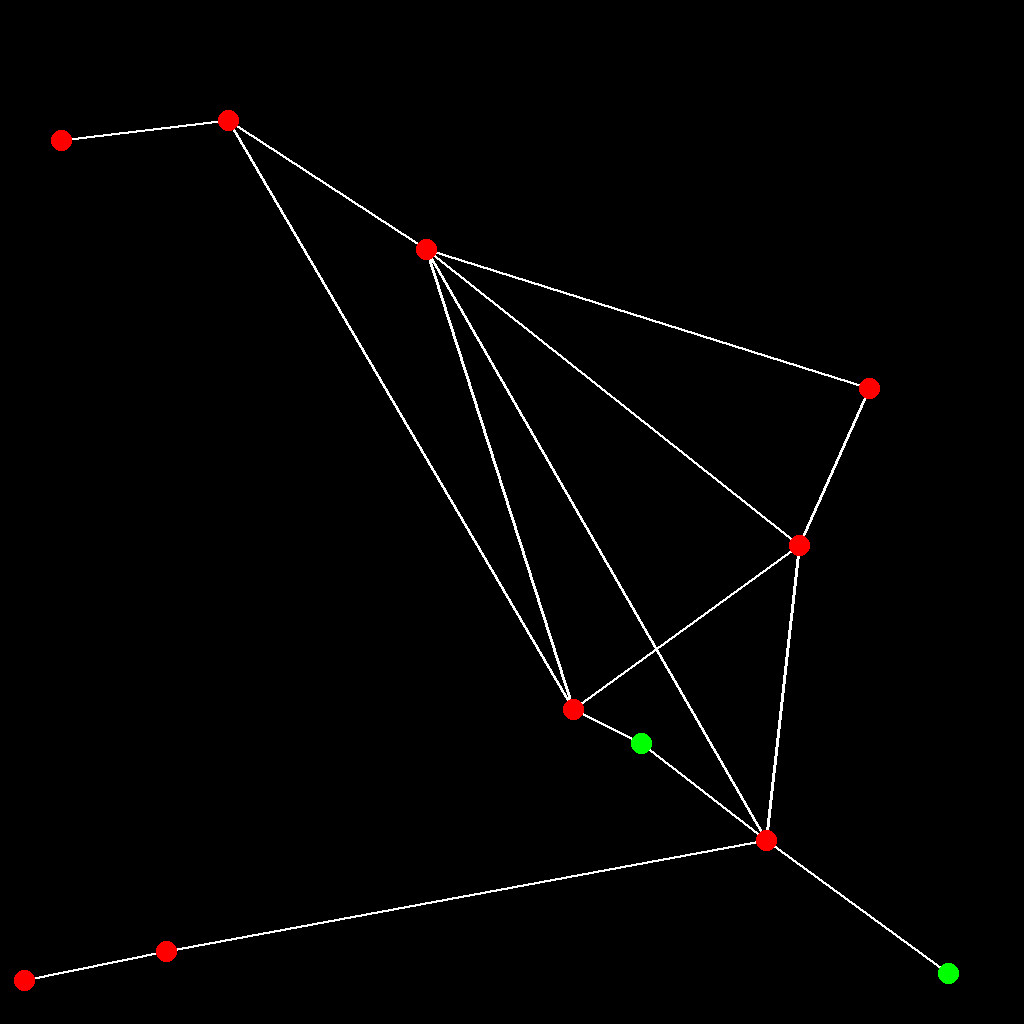} &
        \includegraphics[width=\x]{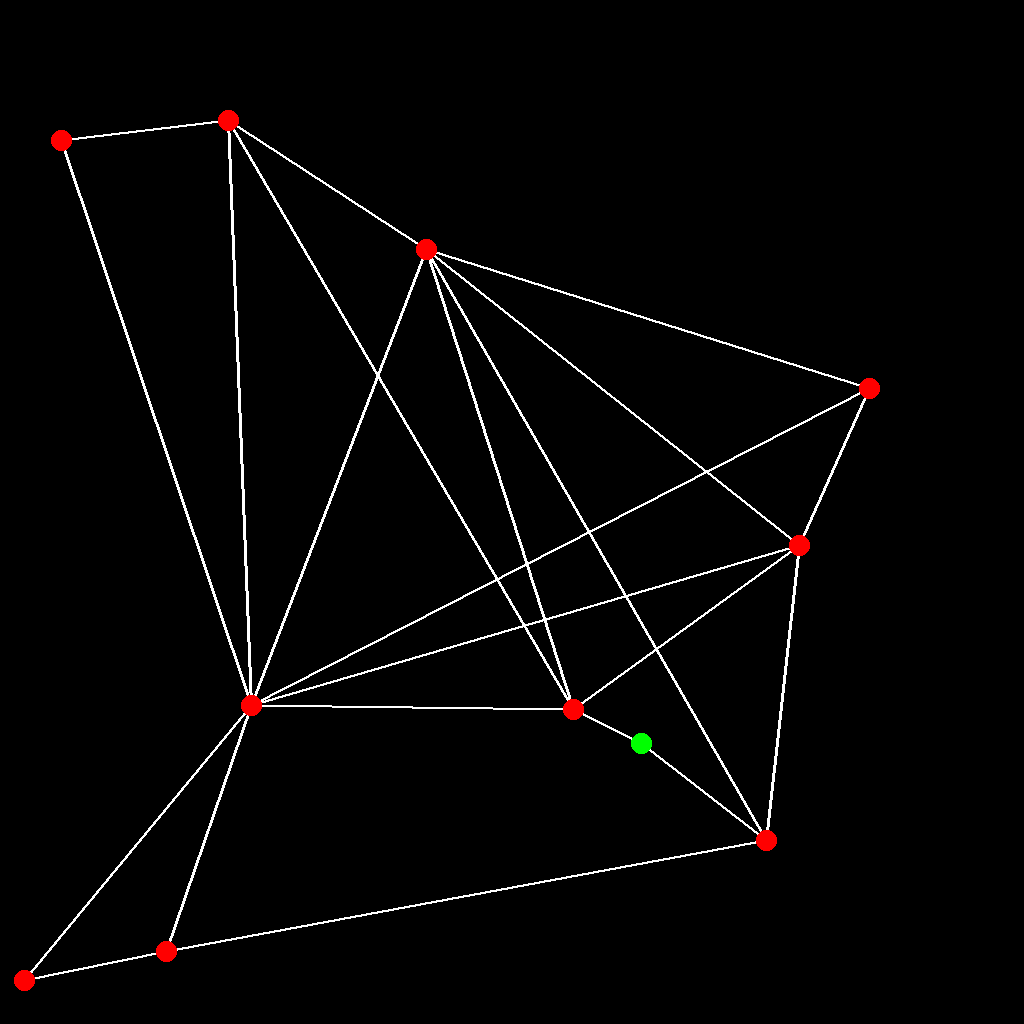} &
        \includegraphics[width=\x]{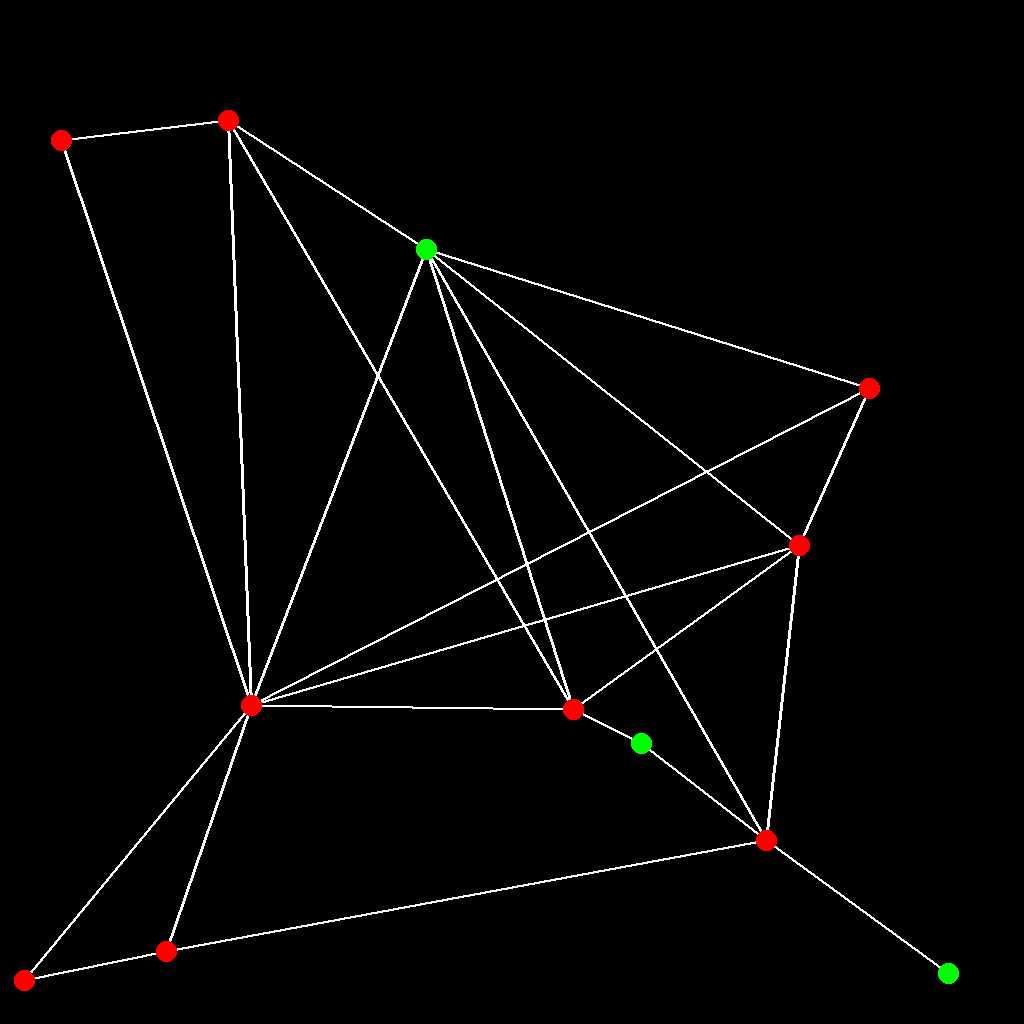} \\
      \rotatebox[origin=l]{90}{\scriptsize{images}} & 
        \includegraphics[width=\x]{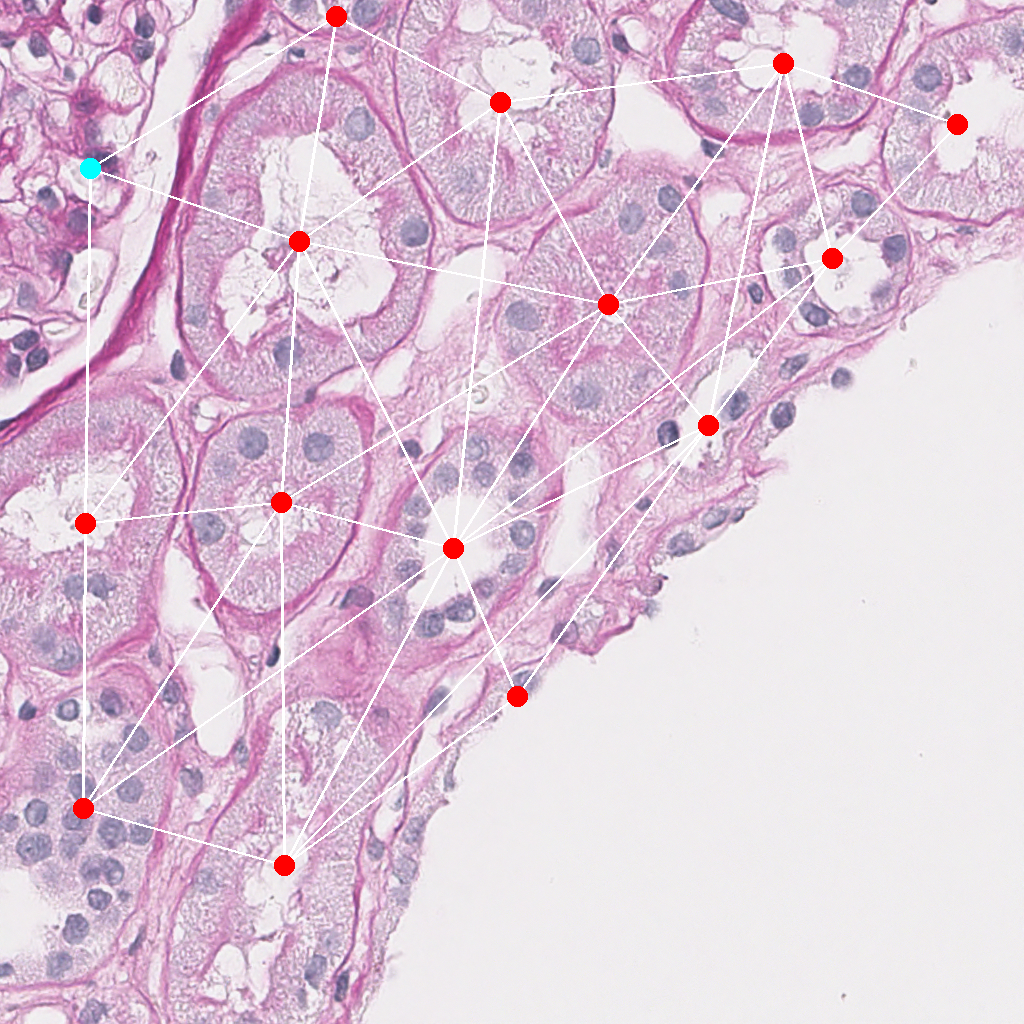} &  
        \includegraphics[width=\x]{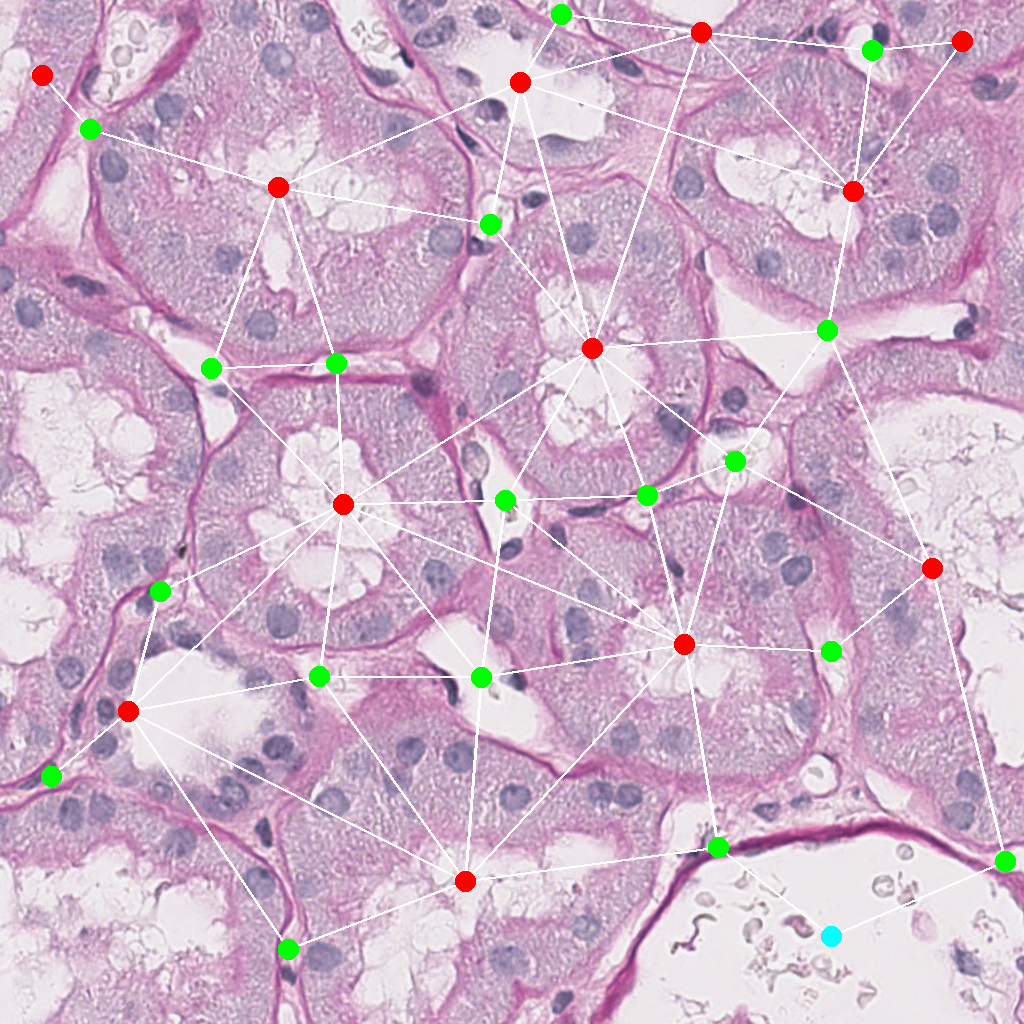} &
        \includegraphics[width=\x]{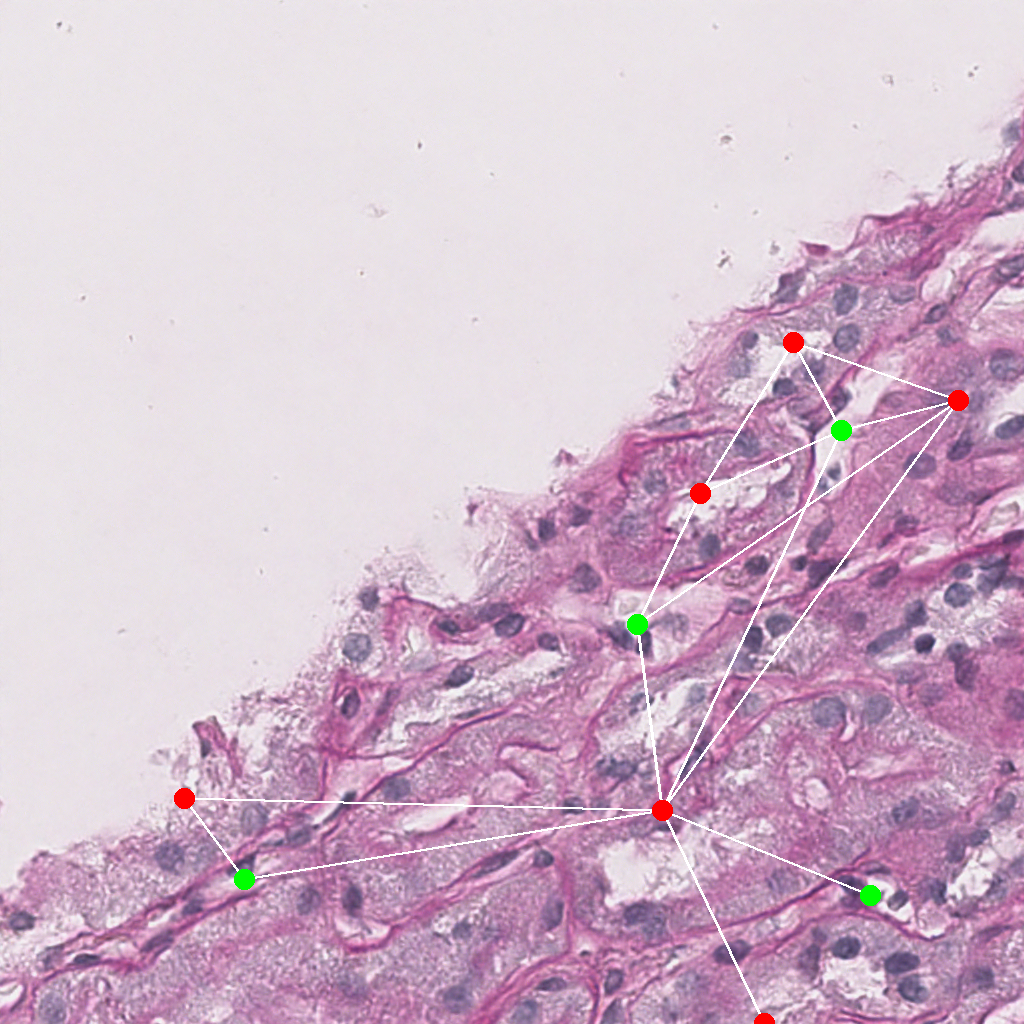} &
        \includegraphics[width=\x]{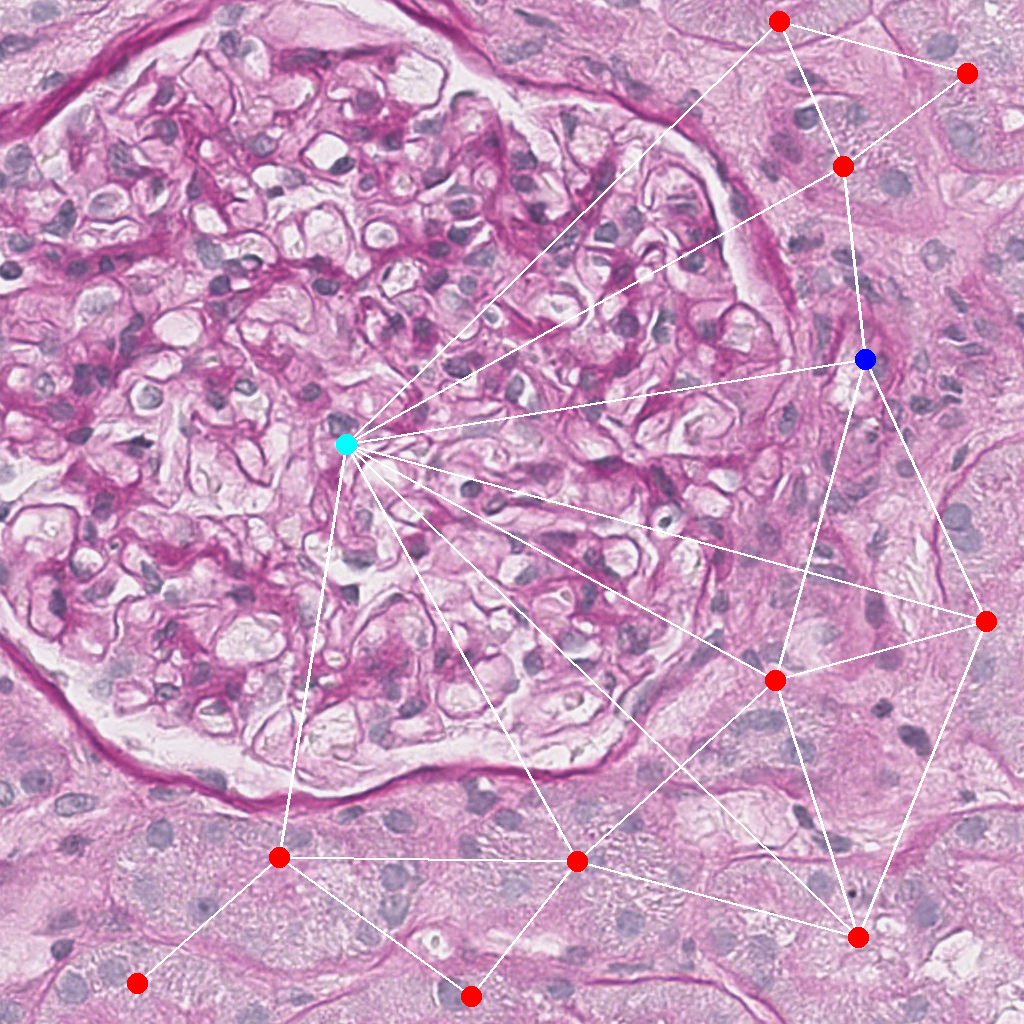} &
        \includegraphics[width=\x]{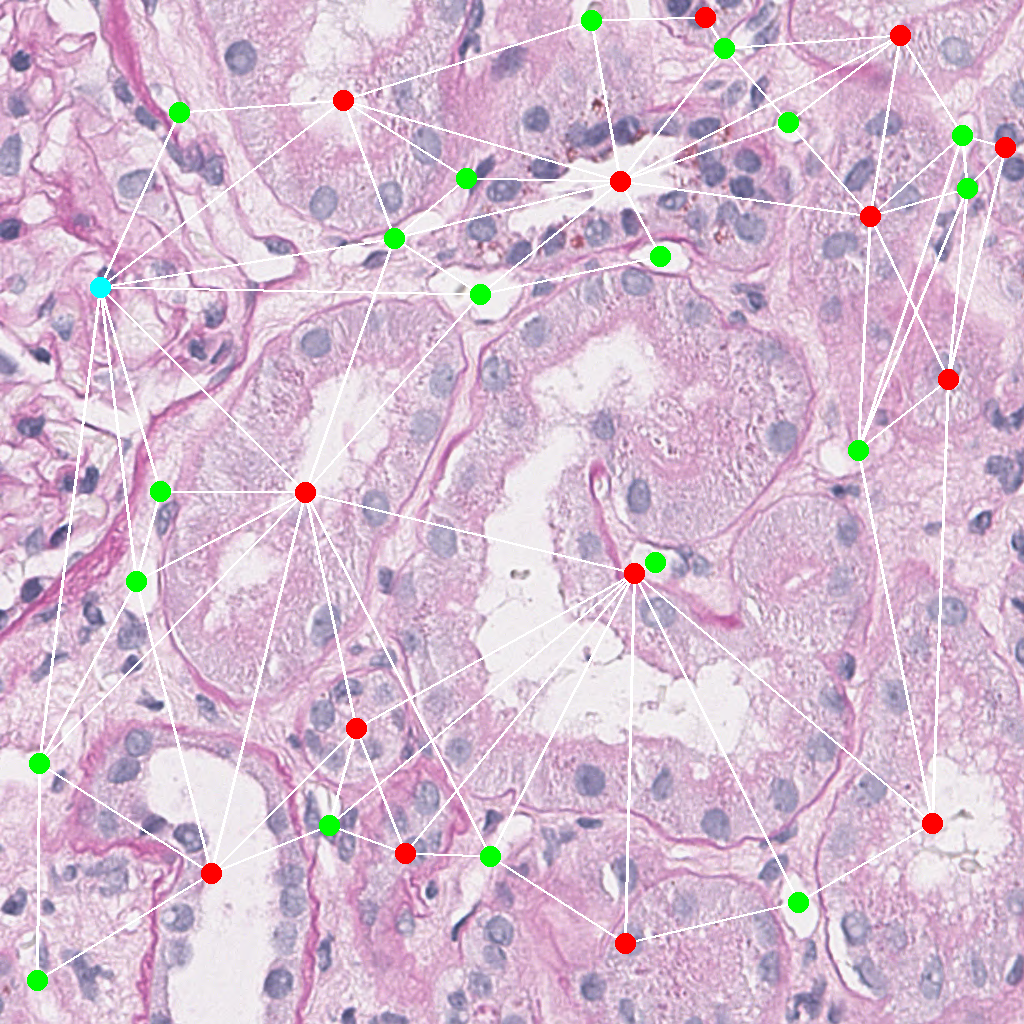} & 
        & 
        \includegraphics[width=\x]{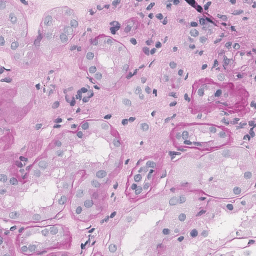} & 
        \includegraphics[width=\x]{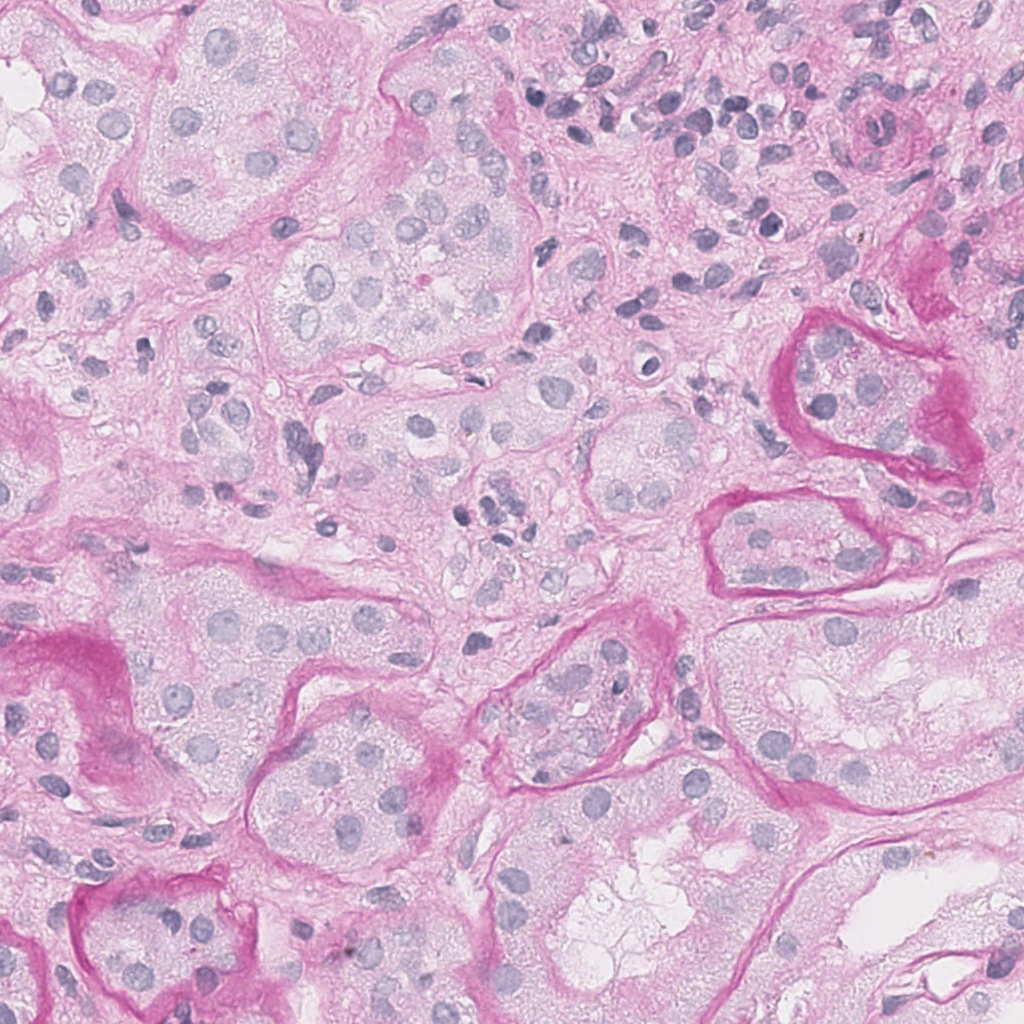} &
        \includegraphics[width=\x]{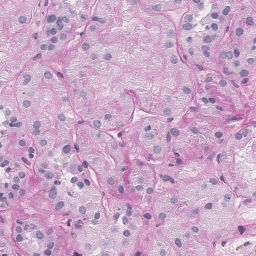} &
        \includegraphics[width=\x]{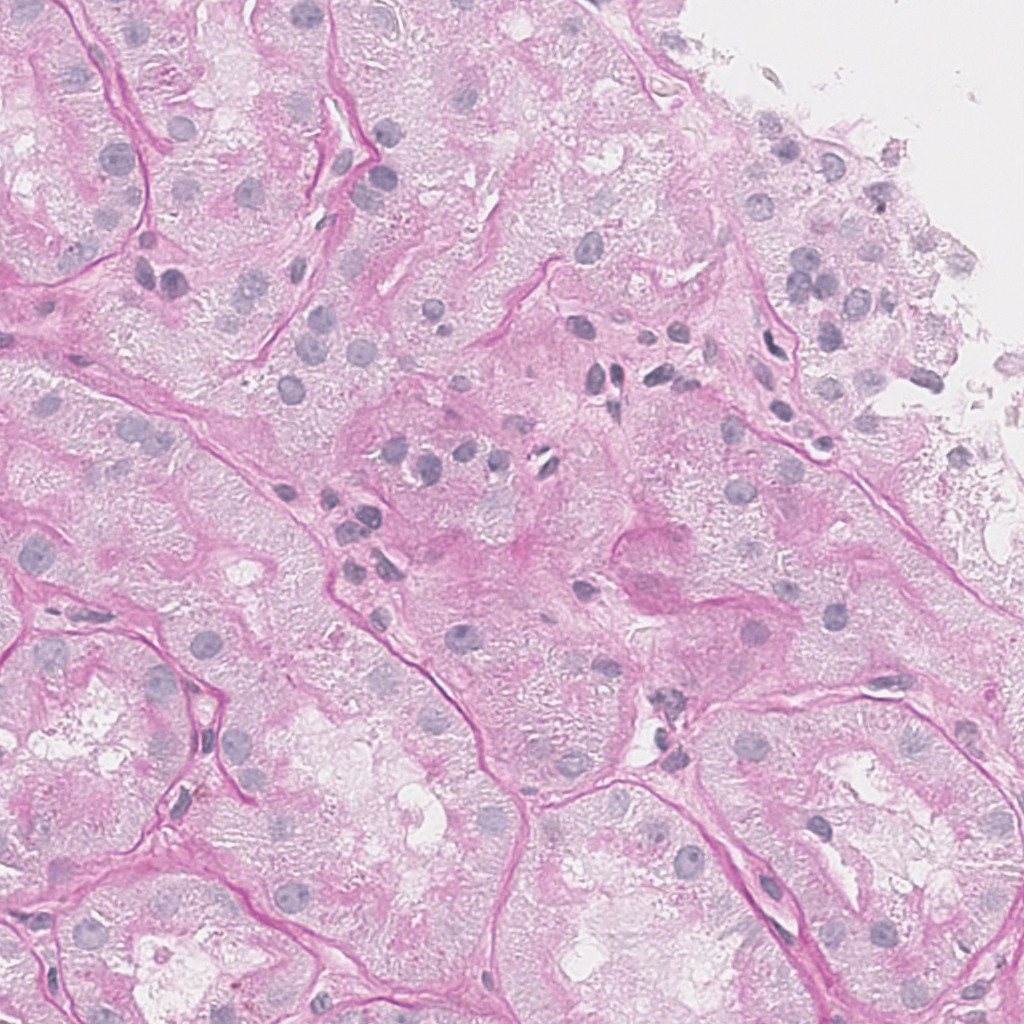} &
        \includegraphics[width=\x]{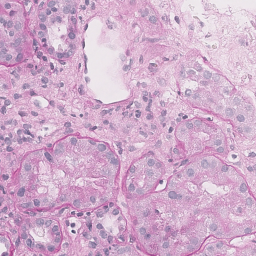} \\
      \rotatebox[origin=l]{90}{\scriptsize{masks}} & 
        \includegraphics[width=\x]{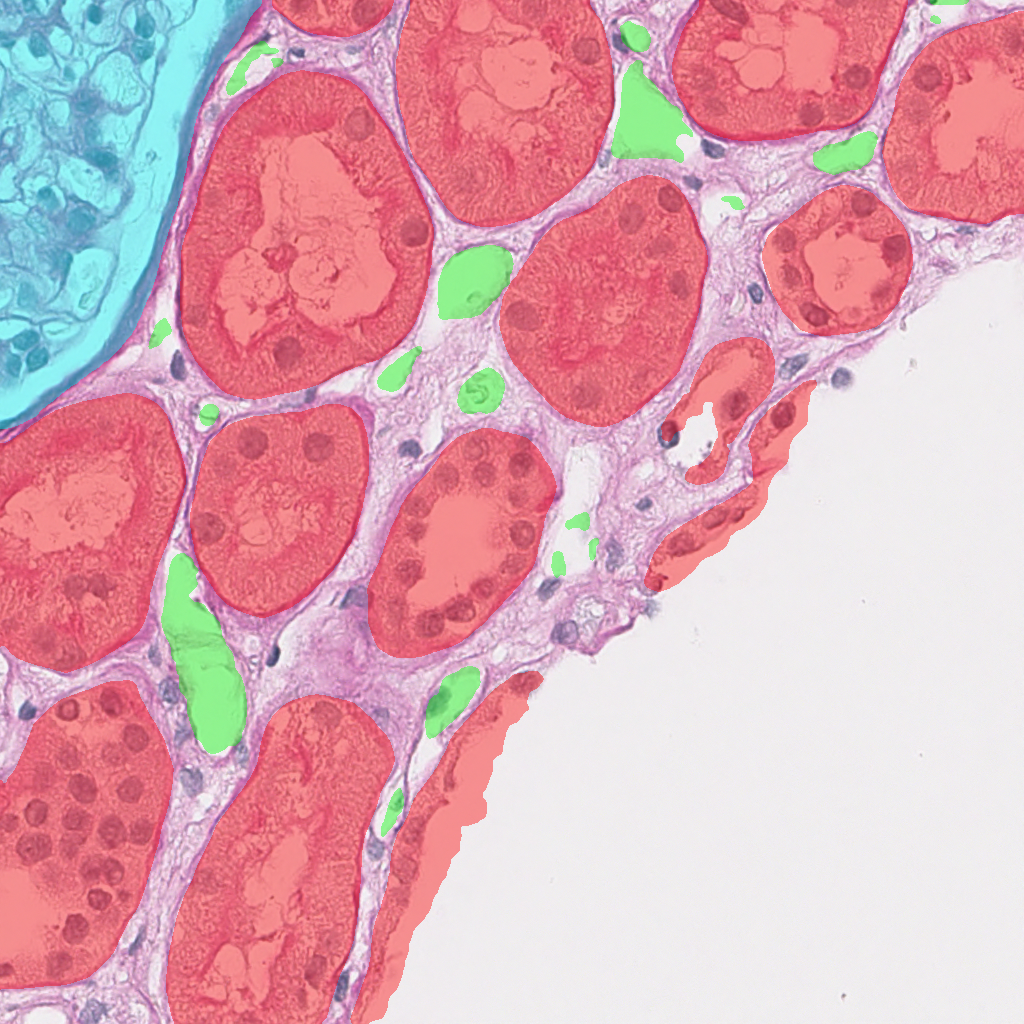} &  
        \includegraphics[width=\x]{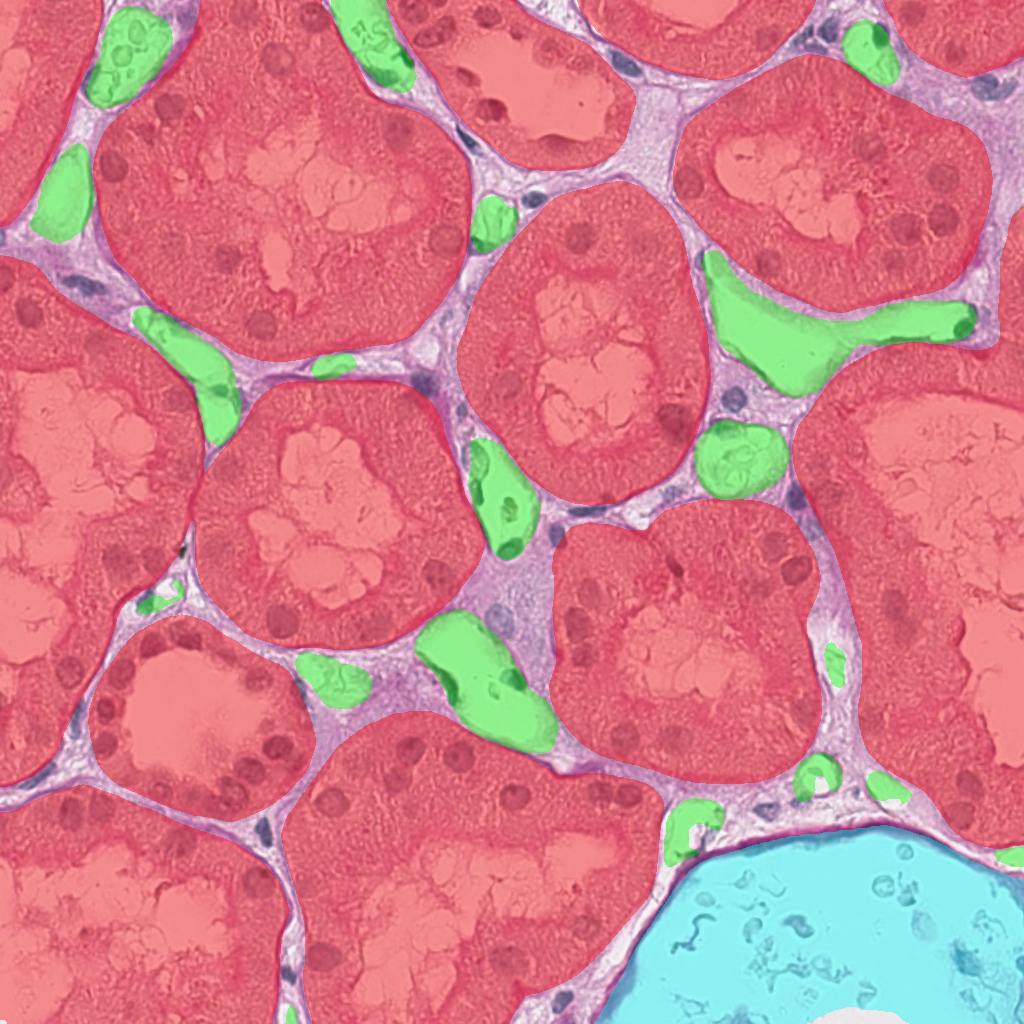} &
        \includegraphics[width=\x]{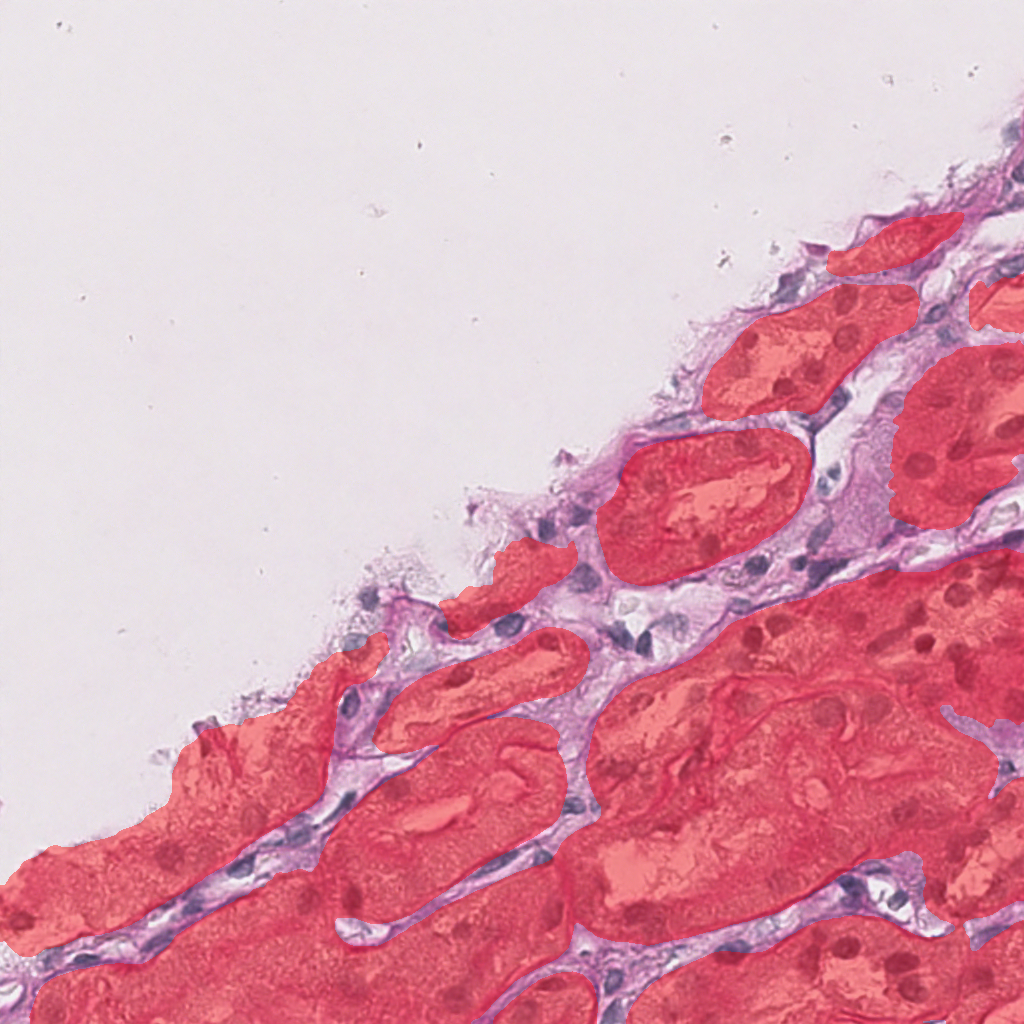} &
        \includegraphics[width=\x]{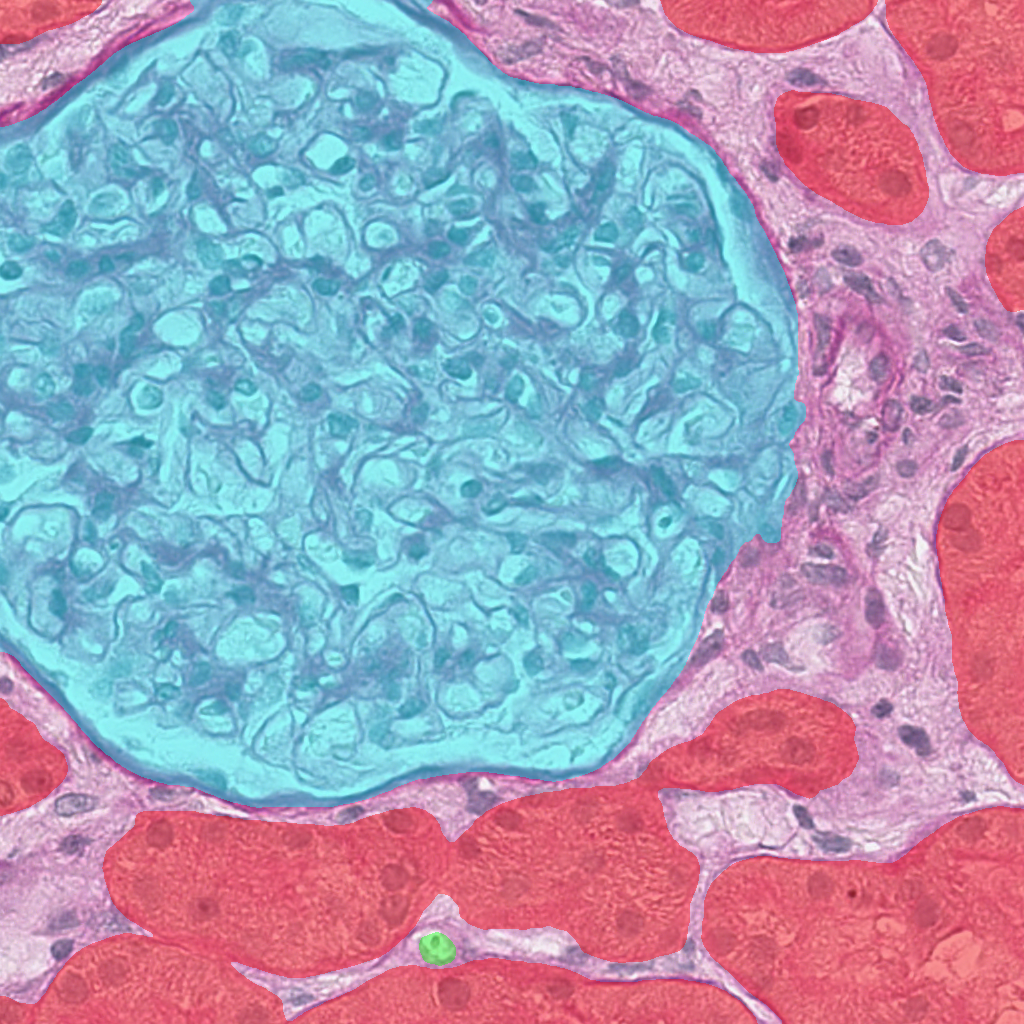} &
        \includegraphics[width=\x]{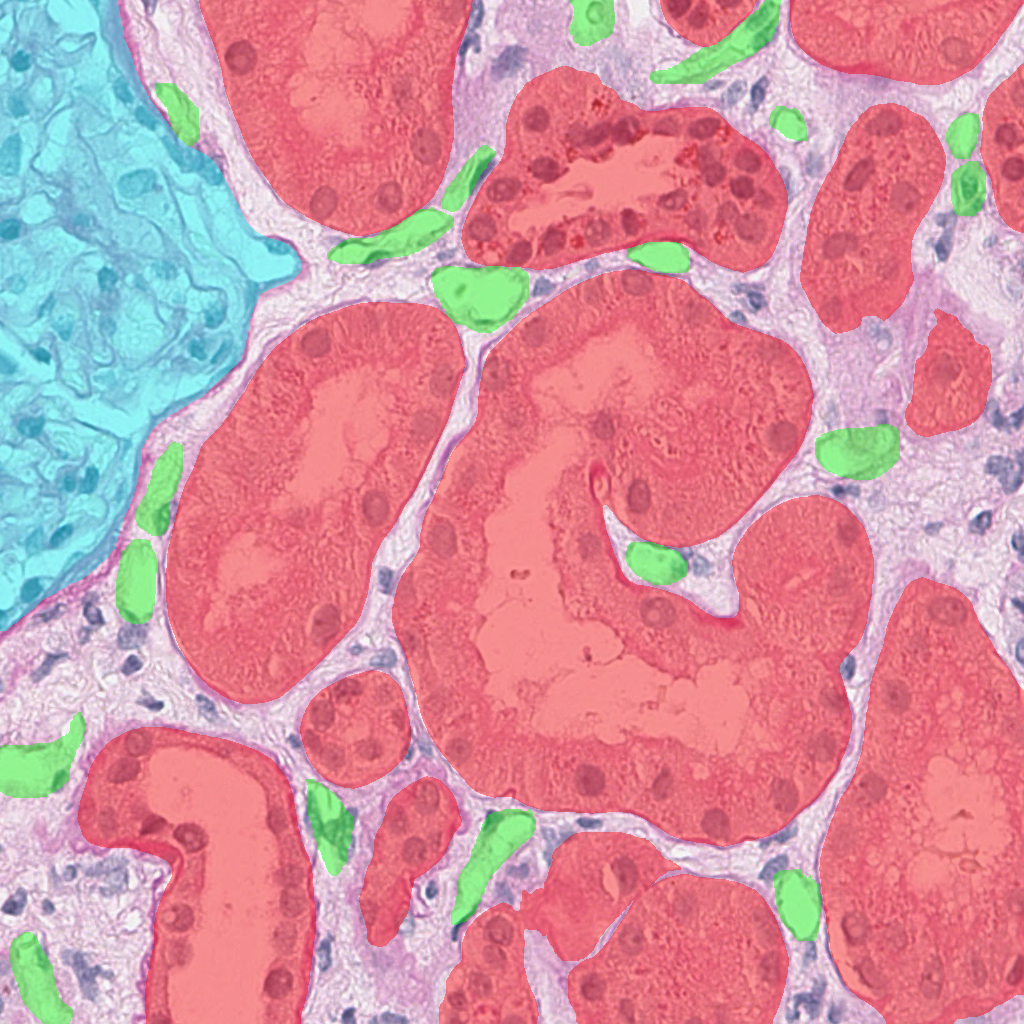} & 
        & 
        \includegraphics[width=\x]{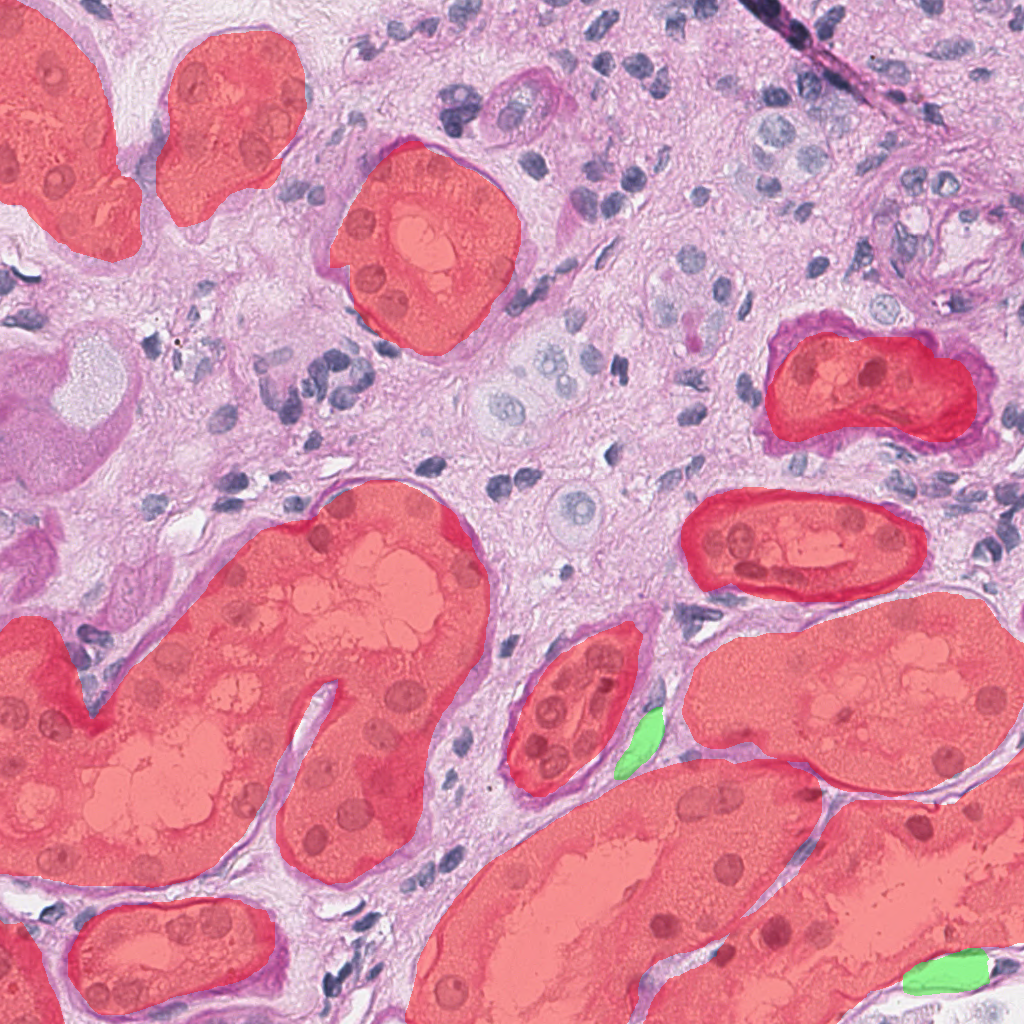} & 
        \includegraphics[width=\x]{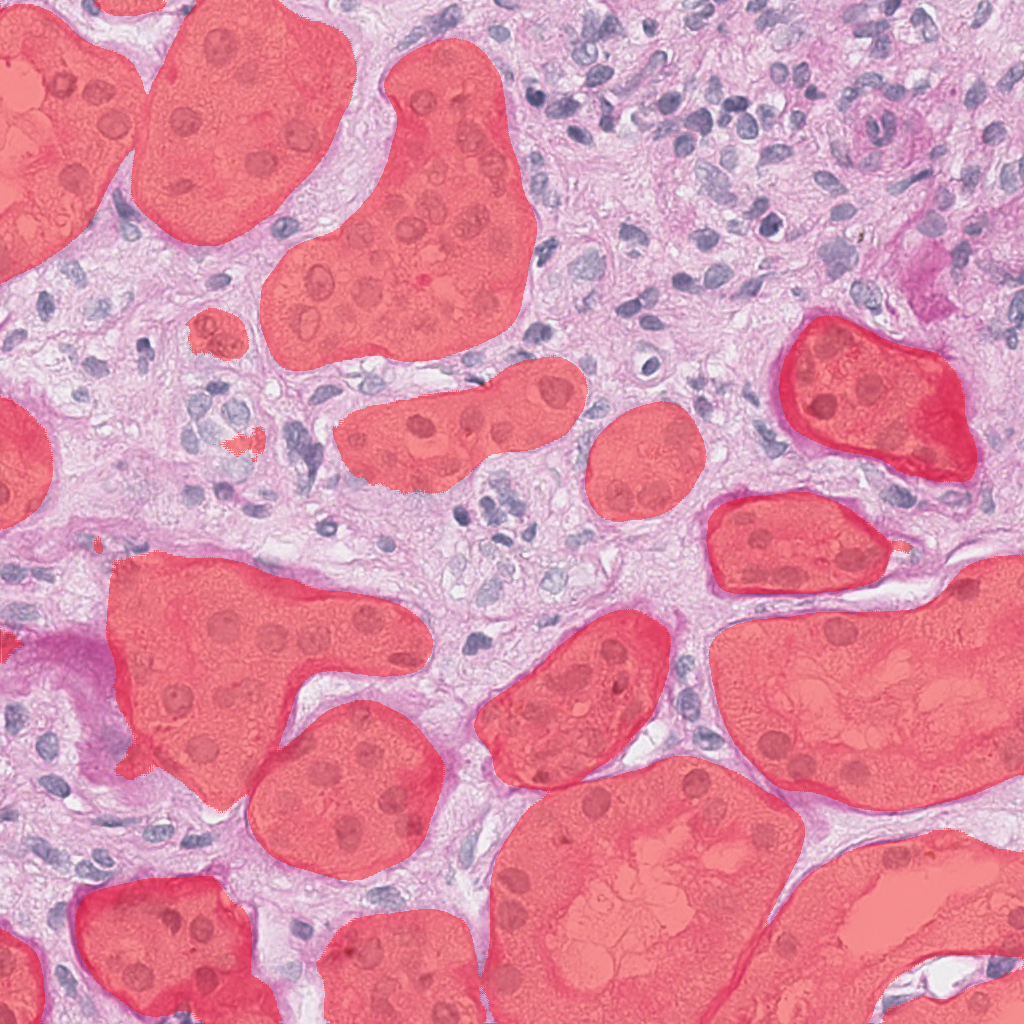} &
        \includegraphics[width=\x]{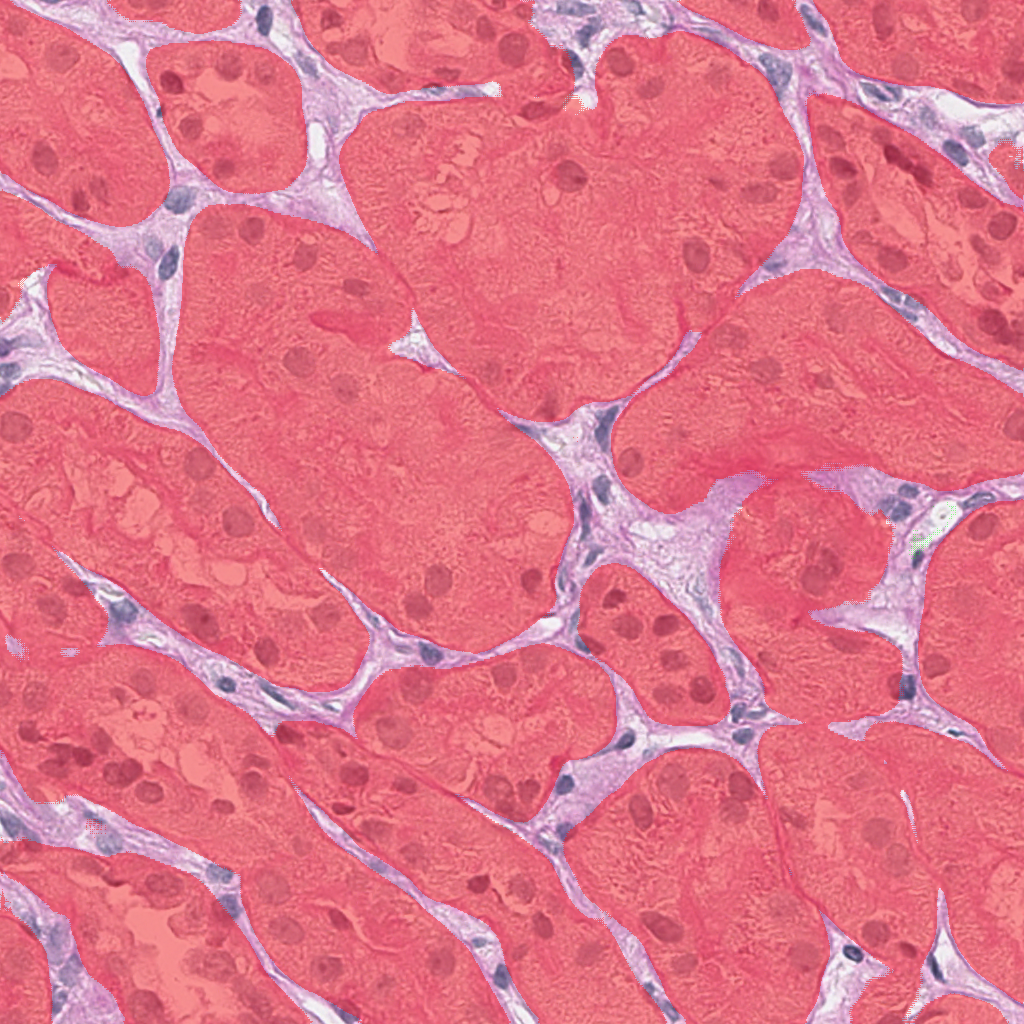} &
        \includegraphics[width=\x]{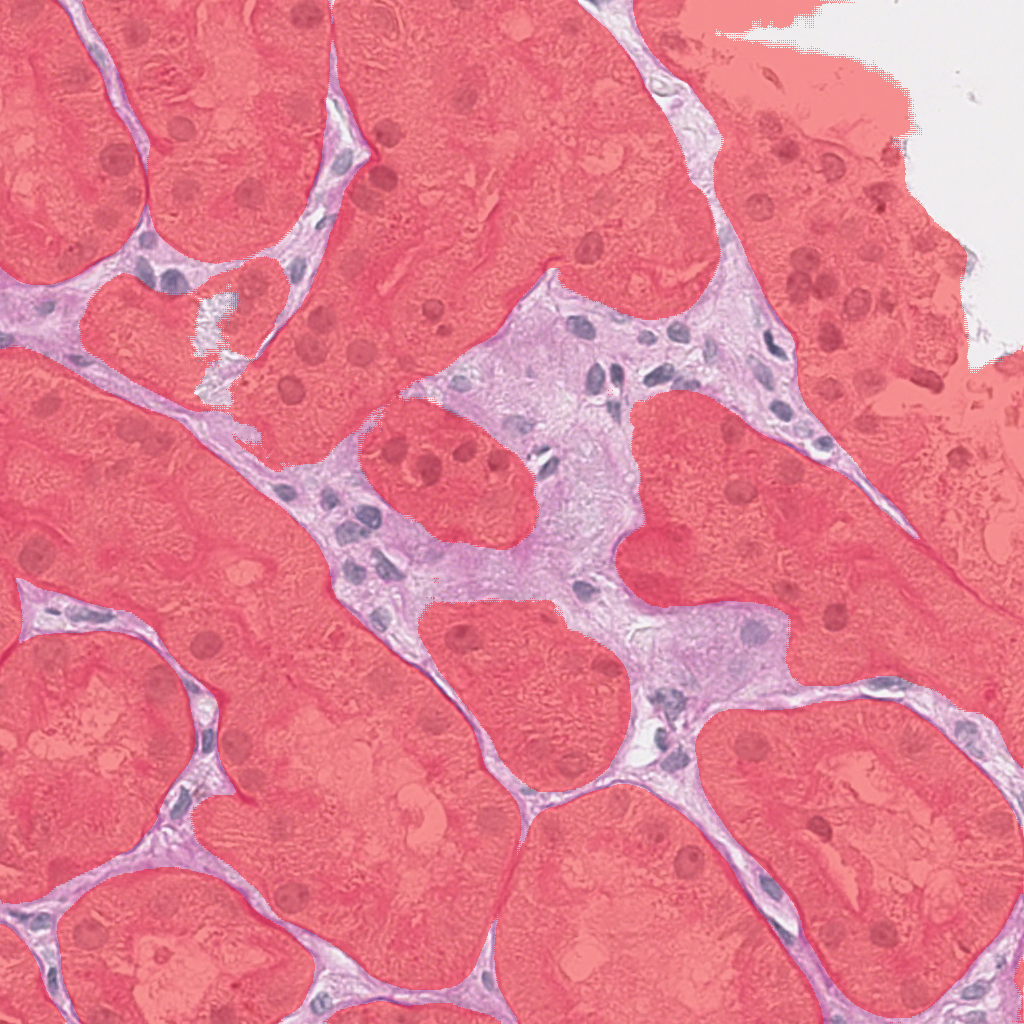} &
        \includegraphics[width=\x]{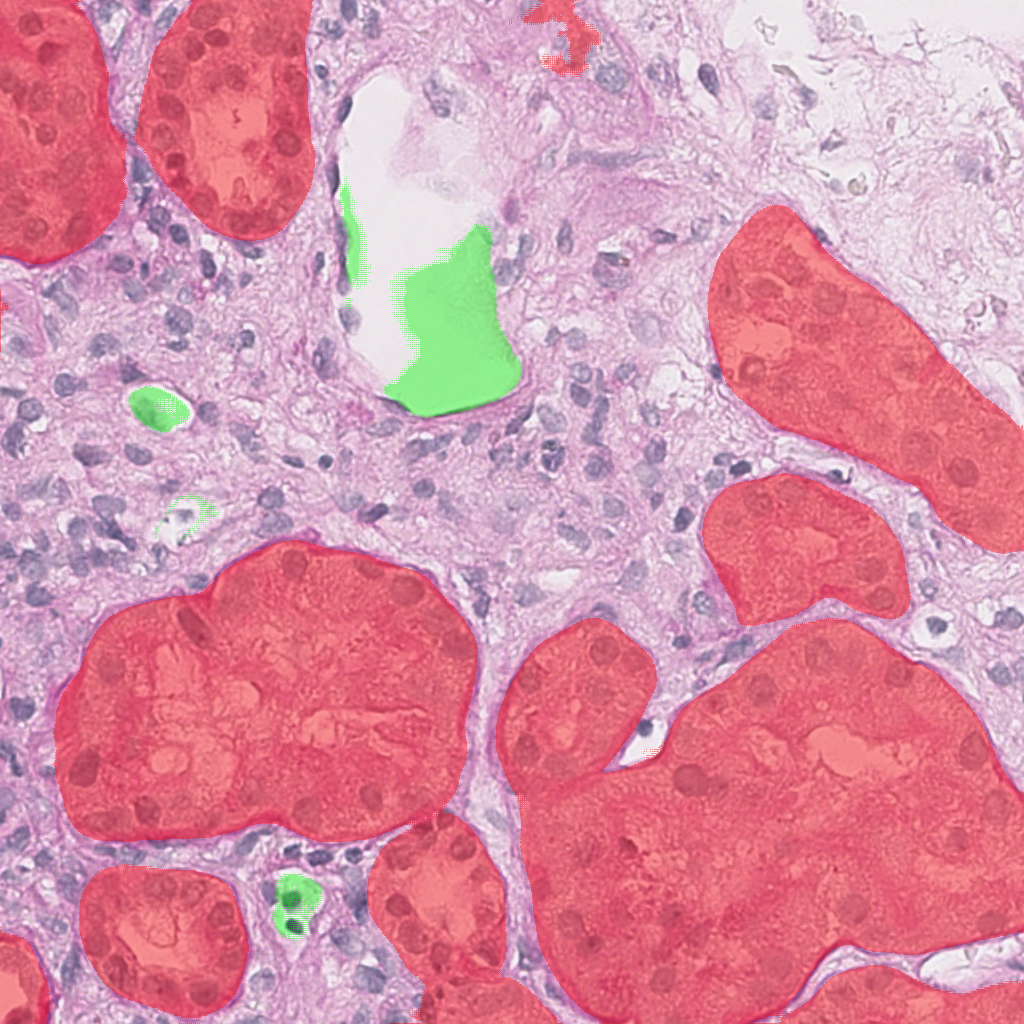} \\
    \end{tabular}
  }
  \caption{Left group: synthetic data and a real example. Right group: causally inspired interventions starting with real graph $G$  and synthetic modifications ($G^{-4}$, $G^{-11}$, $G^{2:1\rightarrow2}$).}
  \label{fig:combined_results}
\end{figure*}
\begin{table}[t]
\centering
\caption{Comparison of dataset diversity and fidelity for various diffusion model methods. Comparison of the original dataset for generated images on downstream segmentation tasks.}
\resizebox{0.9\textwidth}{!}{ 
\begin{tabular}{llll!{\vrule width 1pt}ll}
\hline
Method & IP$\uparrow$ & IR$\uparrow$ & FID$\downarrow$ & Dice(\%)$\uparrow$ & AJI(\%)$\uparrow$ \\
\hline
Original images & n.a & n.a & n.a & 88.01 & 62.05 \\
\hline
Unconditional Diffusion & 0.82 & 0.57 & \textbf{10.35} & \textbf{90.44} & \textbf{66.80} \\
Mask Conditioned Diffusion & 0.36 & 0.07 & 162.43 & 82.00 & 42.40 \\
\hline
Graph Image Conditioned Diffusion & \textbf{0.90} & 0.30 & 79.11 & 89.85 & 66.60 \\
Graph Text Conditioned Diffusion & 0.77 & \textbf{0.64} & 39.78 & 86.05 & 59.45 \\
\hline
\end{tabular}
}
\label{tab:methodcomparison}
\end{table}

\noindent\textbf{Ablation Study:} 
To understand the effect of each graph modification, we conducted an ablation study examining different combinations of graph generation and graph conditioning methods, as seen in Table \ref{tab:ablation}. Further, an example of changes for an individual graph is illustrated in Figure~\ref{fig:combined_results} right. 
\begin{table}[t]
\centering
\caption{Different methods of feature extractions used as node vector encodings.
The method names are the graph intervention followed by the graph encoding. `Real' uses real graphs from the train set, `Cut-Paste' and `Cut-Paste(short)' are examples of $G_{new} = G_i \oplus G_j$ where short restricts the number of subgraphs to 2, and `Interpolated' linearly interpolates between two graphs.
`Image' stands for Graph-Image-Conditioned-Model, `Extracted' follows the graph generation models using $
\mathbf{F} = \mathbf{E}_{class}  \oplus \mathbf{E}_{BYOL} \oplus \mathbf{E}_{pos}, 
$ and `Manual' relies on manual graph extraction using a centroid, average areas and bounding boxes for each node.}
\begin{adjustbox}{valign=c, width=0.9\textwidth} 
\setlength{\tabcolsep}{3pt} 
\begin{tabular}{llll!{\vrule width 1pt}ll}
\toprule
Method & IP$\uparrow$ & IR$\uparrow$ & FID$\downarrow$ & Dice(\%)$\uparrow$ & AJI(\%)$\uparrow$\\
\midrule
Real + Manual &0.08&0.00 &322.7 & 21.59&14.11\\
Real + Extracted &0.77&0.64 &\textbf{39.78} &86.05 &59.45\\
Real + Image&0.81&0.49 & 55.65 & 87.35&61.48\\
Cut-Paste + Manual &0.05&0.00 & 326.8 & 33.64 &16.93\\
Cut-Paste + Extracted &0.70&0.60 & 186.35 &37.18&22.32 \\
Cut-Paste + Image &0.88&0.41 & 83.19 &82.68&52.73\\
Cut-Paste(short) + Manual &0.10&0.00 & 326.6 & 87.39& 61.59 \\
Cut-Paste(short) + Extracted &0.71&\textbf{0.75} & 199.2 & 58.87& 31.08\\
Cut-Paste(short) + Image &0.89&0.37 & 77.93 & 79.27 & 48.36 \\
Interpolated + Manual &0.10&0.00 & 327.7 & 36.20 &21.34 \\
Interpolated + Extracted &0.65&0.35 & 201.1 &89.35&65.31 \\
Interpolated + Image &\textbf{0.90}&0.30 & 79.11 &\textbf{89.85}&\textbf{66.60} \\
\bottomrule
\end{tabular}
\end{adjustbox}
\label{tab:ablation}
\end{table}

\noindent\textbf{Evaluation:}
As shown in \Cref{tab:methodcomparison,tab:ablation} and \Cref{fig:combined_results}, the GCD models perform effectively across various tasks, consistently producing images that faithfully represent the graphs they are conditioned on both in image and textual form. They generate a more diverse set of samples compared to other methods while maintaining high accuracy, as evidenced by the IP and IR scores in Table \ref{tab:methodcomparison}.
Interestingly, occasionally less faithful images that are noisier in appearance lead to higher Dice and AJI scores, as shown in Table~\ref{tab:ablation}, on downstream tasks.
This finding suggests that diversity in generated samples may be more critical for downstream tasks than the realism of individual images.
The ablation study results (Table \ref{tab:ablation}) 
indicate that linear interpolations may be more effective than more complex structural manipulations, such as Cut-Paste.
This is likely due to the individual node representations already capturing critical information encoded in the $\mathbf{E}_{BYOL}$, $\mathbf{E}_{pos}$, or the adjacency matrix $\mathbf{A}$, which in turn may result in conflicting signals being transmitted to the model. This is further underlined by simpler changes in  Cut-Paste(short) giving better results then  Cut-Paste. It is important to note that all these manipulations were used as replacements to train new segmentation models, rather than as data augmentations.
As a result, the Dice scores reflect performance on purely synthetic images, not augmented ones. However, as per \cite{Cechnicka2023}, strong synthetic performance often leads to improved results when combining synthetic and real data.

\noindent\textbf{Discussion:}
The results show that simple, rule-based interventions lead to better outcomes than more complex transformations. Although this may seem counter-intuitive, it aligns with the sparse representation theory~\cite{elad2006image} in image reconstruction, which suggests that regularized, simplified structures capture essential features more effectively, while excessive complexity can disrupt these features. The GCD model can be seen as a `sparsity-guided diffusion' paradigm, where its interventions focus on connectivity and relational patterns, generating clinically relevant diversity without introducing unnecessary complexity.
\section{Conclusion}


In this work, we introduced a novel approach to generating synthetic histopathological images by introducing GCD models. Our approach leverages graph-based representations to enhance diversity, fidelity, and control in synthetic image generation. By explicitly encoding spatial relationships and anatomical structures, our method preserves critical structural consistency, addressing longstanding challenges in synthetic medical imaging. Furthermore, targeted interventions enrich dataset diversity in clinically meaningful ways. 
Our results demonstrate that GCD not only improves image diversity metrics but also achieves comparable or superior performance in downstream segmentation tasks compared to traditional methods.

\begin{credits}
\subsubsection{\ackname} S. Cechnicka is supported by the UKRI Centre for Doctoral Training AI4Health  (EP / S023283/1). Support was also received from the ERC project MIA-NORMAL 101083647, the State of Bavaria (HTA) and DFG 512819079. HPC resources were provided by NHR@FAU of FAU Erlangen-N\"urnberg under the NHR project b180dc. NHR@FAU hardware is partially funded by the DFG – 440719683.
Dr. Roufosse is supported by the National Institute for Health Research (NIHR) Biomedical Research Centre based at Imperial College Healthcare NHS Trust and Imperial College London (ICL). The views expressed are those of the authors and not necessarily those of the NHS, the NIHR or the Department of Health. Dr Roufosse’s research activity is made possible with generous support from Sidharth and Indira Burman.
Human samples used in this research project were obtained from the Imperial College Healthcare Tissue \& Biobank (ICHTB). ICHTB is supported by NIHR Biomedical Research Centre based at Imperial College Healthcare NHS Trust and ICL. ICHTB is approved by Wales REC3 to release human material for research (22/WA/2836)
\subsubsection{\discintname}
The authors have no competing interests to declare that are relevant to the content of this article.
\end{credits}
\newpage
\bibliographystyle{splncs04}
\bibliography{samplebibliography}

\end{document}